\documentclass[runningheads]{llncs}

\usepackage{eccv}

\usepackage{eccvabbrv}

\usepackage{graphicx}
\usepackage{enumitem}
\usepackage{booktabs}
\usepackage[ruled,vlined]{algorithm2e}
\usepackage[dvipsnames, table]{xcolor}
\usepackage[hang,flushmargin]{footmisc}
\usepackage{pifont}

\usepackage{soul}
\sethlcolor{yellow}

\usepackage[accsupp]{axessibility}  

\usepackage{hyperref}

\usepackage{orcidlink}

\begin{document}

\title{Stream-DiffVSR: Low-Latency Streamable Video Super-Resolution via Auto-Regressive Diffusion} 

\titlerunning{Stream-DiffVSR}

\author{Hau-Shiang Shiu\inst{1} \and Chin-Yang Lin\inst{1} \and Zhixiang Wang\inst{2} \and Chi-Wei Hsiao\inst{3} \and Po-Fan Yu\inst{1} \and Yu-Chih Chen\inst{1} \and Yu-Lun Liu\inst{1}}

\authorrunning{H.-S.~Shiu et al.}

\institute{\textsuperscript{\rm 1} Department of Computer Science, National Yang Ming Chiao Tung University,\\\textsuperscript{\rm 2} Shanda AI Research Tokyo, \textsuperscript{\rm 3} MediaTek Inc.\\
\email{xhs0964519.cs12@nycu.edu.tw, yulunliu@cs.nycu.edu.tw}}

\maketitle

\renewcommand\twocolumn[1][]{#1}%
\begin{center}
\centering
\captionsetup{type=figure}
\vspace{-3mm}
\resizebox{1.0\textwidth}{!} 
{
\includegraphics[width=\textwidth]{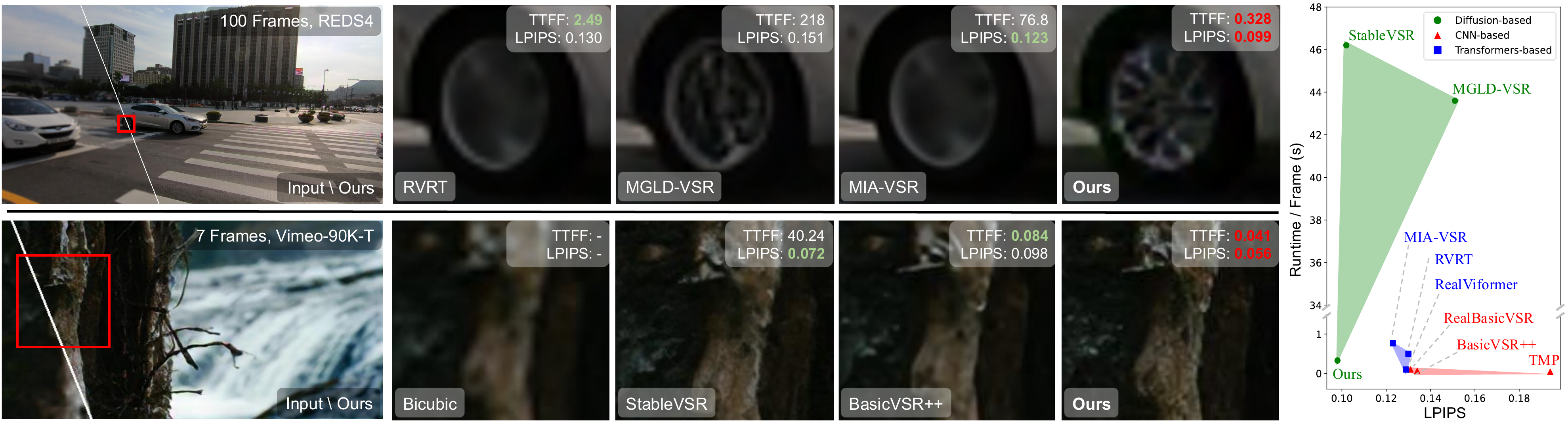}
}
\vspace{-6mm}
\caption{\textbf{Comparison of visual quality and inference speed across VSR methods.} Stream-DiffVSR achieves superior perceptual quality and comparable runtime to CNN- and Transformer-based models, with significantly reduced Time-to-First-Frame (TTFF) versus offline approaches. Best/second-best in \textbf{\textcolor{red}{red}}/\textbf{\textcolor{Green}{green}}.}
\label{fig:teaser}
\end{center}

\begin{abstract}
Diffusion-based video super-resolution (VSR) methods deliver strong perceptual quality but are often unsuitable for latency-sensitive scenarios due to reliance on future frames and expensive multi-step denoising. We propose Stream-DiffVSR, a causally conditioned diffusion framework for efficient online VSR. Operating strictly on past frames, Stream-DiffVSR integrates a four-step distilled denoiser for fast inference, an Auto-regressive Temporal Guidance (ARTG) module that injects motion-aligned cues during latent denoising, and a lightweight temporal-aware decoder with a Temporal Processor Module (TPM) to enhance detail and temporal coherence. Unlike chunk-wise streaming inference, our strictly frame-by-frame causal design avoids sequence-level waiting, substantially reducing time-to-first-frame and end-to-end latency. Stream-DiffVSR processes 720p frames in 0.328 seconds on an RTX 4090 and consistently outperforms prior diffusion-based baselines. Compared with the online state-of-the-art TMP~\cite{zhang2024tmp}, it improves perceptual quality (LPIPS +0.095). Compared with prior diffusion-based VSR methods such as MGLD-VSR~\cite{yang2024motion}, it reduces per-frame runtime by over 130×. Moreover, Stream-DiffVSR substantially lowers time-to-first-frame for diffusion-based VSR, reducing initial delay from over 4600 seconds to 0.328 seconds, making diffusion-based VSR markedly more practical for low-latency online and streaming deployment.
Project page: \url{https://jamichss.github.io/stream-diffvsr-project-page/}
\keywords{Video super-resolution \and Diffusion models \and Online/Streaming inference}
\end{abstract}    
\section{Introduction}
\label{sec:intro}
Video super-resolution (VSR) aims to reconstruct high-resolution (HR) videos from low-resolution (LR) inputs and is vital in applications such as surveillance, live broadcasting, video conferencing, autonomous driving, and drone imaging. It is increasingly important in low-latency rendering workflows, including neural rendering and resolution upscaling in game engines and AR/VR systems, where latency-aware processing is crucial for visual continuity.

Specifically, latency-sensitive processing involves two key aspects: per-frame inference time (throughput) and end-to-end system latency (delay between receiving an input frame and producing its output). Existing VSR methods often struggle with this trade-off. While CNN- and Transformer-based models offer a balance between efficiency and quality, they fall short in perceptual detail. Diffusion-based models excel in perceptual quality due to strong generative priors, but suffer from high computational cost and reliance on future frames, making them impractical for time-sensitive video applications.

In this paper, we propose \textbf{Stream-DiffVSR}, a diffusion-based framework that establishes a fully causal, frame-by-frame paradigm for online video super-resolution, bridging the gap between high-quality but slow diffusion methods and fast but lower-quality CNN- or Transformer-based methods. Unlike previous diffusion-based VSR approaches (e.g., StableVSR~\cite{rota2024enhancing} and MGLD-VSR~\cite{yang2024motion}) that typically require 50 or more denoising steps and bidirectional temporal information, we adopt a staged training scheme with rollout distillation over full inference trajectories, reducing denoising steps to just four while remaining robust under autoregressive rollout. Additionally, we introduce an Auto-regressive Temporal Guidance mechanism and an Auto-regressive Temporal-aware Decoder, with a learnable gating parameter controlling reliance on the previous frame, to effectively exploit temporal information from previous frames, significantly enhancing temporal consistency and perceptual fidelity.

\cref{fig:teaser} illustrates the core advantage of our approach by comparing visual quality and runtime across various categories of video super-resolution methods. Stream-DiffVSR achieves improved perceptual quality, as reflected by lower LPIPS~\cite{zhang2018unreasonable} and temporal consistency, outperforming existing unidirectional CNN- and Transformer-based methods (e.g., MIA-VSR~\cite{zhou2024video}, RealViformer~\cite{zhang2024realviformer}, TMP~\cite{zhang2024tmp}). Notably, Stream-DiffVSR offers significantly faster per-frame inference than prior diffusion-based approaches (e.g., StableVSR~\cite{rota2024enhancing}, MGLD-VSR~\cite{yang2024motion}), attributed to our use of a distilled 4-step denoising process and a lightweight temporal-aware decoder.

In addition, existing diffusion-based methods, such as StableVSR~\cite{rota2024enhancing} typically rely on bidirectional or future-frame information, resulting in prohibitively high processing latency that is not suitable for online scenarios. Specifically, for a 100-frame video, StableVSR (46.2 s/frame) would incur an initial latency exceeding 4600 seconds on an RTX 4090 GPU, as it requires processing the entire sequence before generating even the first output frame. In contrast, our Stream-DiffVSR operates in a strictly causal, autoregressive manner, conditioning only on the immediately preceding frame. Consequently, the initial frame latency of Stream-DiffVSR corresponds to a single frame's inference time (0.328 s/frame), reducing the latency by more than three orders of magnitude compared to StableVSR. This significant latency reduction demonstrates that Stream-DiffVSR effectively unlocks the potential of diffusion models for practical, low-latency online video super-resolution.


\newcommand{\marksize}{\LARGE}
\newcommand{\cmark}{\textcolor{Green}{\large\ding{51}}}%
\newcommand{\xmark}{\textcolor{red}{\large\ding{55}}}%
\newcommand{\maybe}{\ensuremath{\marksize\triangle}}%
\begin{table}[t]
\centering
\caption{
\textbf{Comparison of diffusion-based VSR methods.} We report online capability, inference steps, runtime (FPS on 720p, RTX 4090), Time-to-First-Frame (TTFF in sec), and whether each method uses distillation, temporal modeling, or offline future frames. OOM denotes out-of-memory; - indicates missing public inference results. Notably, Stream-DiffVSR runs in a strictly online, past-only setting and achieves one of the lowest end-to-end latencies among compared diffusion-based VSR methods under our measurement protocol.}
\label{tab:vsr_comparison}
    \vspace{-3mm}
\centering
\small
\setlength{\tabcolsep}{3pt}
\resizebox{\columnwidth}{!}{%
\begin{tabular}{lccccccc}
\toprule
& & {\# of} & {FPS} & {} & & {Temporal} & {Temporal} \\
{Method} & {Online} & {Steps} & {@720p} & {TTFF (sec)}& {Distill} & {Input} & {Decoder} \\
\midrule
StableVSR~\cite{rota2024enhancing}     & \xmark & 50  & 0.02 & 4620 & \xmark & Future/Bi-dir & \xmark \\
MGLD-VSR~\cite{yang2024motion}         & \xmark & 50  & 0.02 & 218 & \xmark & Future/Bi-dir & \cmark \\
Upscale-A-Video~\cite{zhou2024upscale} & \xmark & 30  & OOM & - & \xmark & Future/Bi-dir & \cmark \\
VEnhancer~\cite{he2024venhancer}       & \xmark & 15  & OOM & - & \xmark & Future/Bi-dir & \cmark \\
\rowcolor{black!10}
\textbf{Stream-DiffVSR (ours)}             & \textbf{\cmark} & \textbf{4}  & \textbf{3.05} & 0.328 & \textbf{\cmark} & \textbf{Past-only} & \textbf{\cmark} \\
\bottomrule
\end{tabular}
}
\end{table}

To summarize, the main contributions of this paper are:
\begin{itemize}[leftmargin=*]
    \item We introduce Stream-DiffVSR, a diffusion-based framework explicitly designed for \textbf{streamable, low-latency} video super-resolution, trained via a staged image-to-video scheme and distilled using full-trajectory rollout distillation, enabling efficient inference by reducing diffusion sampling from 50 denoising steps down to 4 steps.
    \item We propose a novel Auto-regressive Temporal Guidance mechanism and a Temporal-aware Decoder to effectively leverage temporal \textbf{cues} \emph{only} from past frames, significantly enhancing perceptual quality and temporal consistency.
    \item Extensive experiments demonstrate that our approach outperforms existing methods on key perceptual and temporal consistency metrics while maintaining \textbf{low-latency inference}, making diffusion-based VSR practical for latency-critical online/streaming and interactive video pipelines.
\end{itemize}

To contextualize our contributions, Table~\ref{tab:vsr_comparison} compares recent diffusion-based VSR methods in terms of online inference capability, runtime efficiency, and temporal modeling. Stream-DiffVSR performs online, low-latency inference in a strictly causal (past-only), frame-by-frame manner, delivering diffusion-level perceptual quality while maintaining strong temporal stability. This substantial latency reduction of over three orders of magnitude compared to prior diffusion-based VSR models makes diffusion-based VSR markedly more practical for latency-critical online applications such as video conferencing and AR/VR.

\section{Related Work}
\label{sec:related}

\paragraph{Video Super-resolution.}
VSR methods reconstruct high-resolution videos from low-resolution inputs through CNN-based approaches~\cite{xue2019video,tian2020tdan,wang2019edvr,chan2021basicvsr,chan2022basicvsr++,sun2025fiper}, deformable convolutions~\cite{tian2020tdan,dai2017deformable,zhu2019deformable}, online processing~\cite{zhang2024tmp,yin2024online,fuoli2023fast}, recurrent architectures~\cite{sajjadi2018frame,fuoli2019efficient,isobe2020video,yi2019progressive,li2020mucan}, flow-guided methods~\cite{zhou2024fmanet,zhang2024gimm,liu2019deep}, Transformer-based models~\cite{vaswani2017attention,liang2022recurrent,liang2022vrt,shi2022rethinking,zhou2024video}, and implicit alignment techniques~\cite{xu2024enhancing}. Causal and streaming paradigms have emerged for low-latency scenarios~\cite{ghasemabadi2024learning,qu2024online,zheng2025efficient}, while structured pruning~\cite{xia2023structured} targets efficient deployment. Despite advances, low-latency online processing with high perceptual quality remains challenging.

\vspace{-.5em}
\paragraph{VSR under Unknown Degradations.}
VSR under unknown degradations studies how to handle diverse and poorly specified degradation processes~\cite{yang2021real,chan2022investigating} through pre-cleaning modules~\cite{chan2022investigating,goodfellow2020generative,wang2021realesrgan,liu2021learning}, online approaches~\cite{zhang2024realviformer}, kernel estimation~\cite{pan2021deep,ji2020real}, synthetic degradations~\cite{jeelani2023expanding,song2024negvsr,zhang2023crafting,chang2020learning}, new benchmarks~\cite{nan2025realisvsr,conde2024aim}, real-time systems~\cite{cao2021egvsr,jiang2025rtsr}, advanced GANs~\cite{chen2024horgan,tsai2023dual,chen2025adversarial}, and Transformer restorers~\cite{zamir2022restormer,liang2021swinir,blau2018perception}. Recent efforts leverage text-to-video priors for VSR with complex, non-ideal degradations~\cite{xie2025star,zhao2025realisvsr}. Warp error-aware consistency~\cite{lei2020blind} emphasizes temporal error regularisation.

\vspace{-.5em}
\paragraph{Diffusion-based Image and Video Restoration.}
Diffusion models provide powerful generative priors~\cite{rombach2021highresolution,esser2021taming,chao2022denoising} for single-image SR~\cite{saharia2022image,li2022srdiff,hsiao2024ref,yue2023resshift,wang2024sinsr}, inpainting~\cite{lugmayr2022repaint,weng2024vires,liu2025corrfill,tsai2025lightsout}, and quality enhancement~\cite{ho2022cascaded,gao2023implicit,wang2024exploiting}. Video diffusion methods include StableVSR~\cite{rota2024enhancing}, MGLD-VSR~\cite{yang2024motion}, DC-VSR~\cite{han2025dcvsr}, DOVE~\cite{zhang2025dove}, UltraVSR~\cite{li2025ultravsr}, Upscale-A-Video~\cite{zhou2024upscale}, DiffVSR~\cite{li2025diffvsr}, DiffIR2VR-Zero~\cite{yeh2024diffir2vr}, VideoGigaGAN~\cite{xu2024videogigagan}, VEnhancer~\cite{he2024venhancer}, temporal coherence~\cite{wang2025temporal}, AVID~\cite{zhang2024avid}, and SeedVR2~\cite{wang2025seedvr2}. Temporal consistency in video diffusion has also been addressed through optical-flow-guided approaches~\cite{liang2024flowvid,yang2024fresco,chu2024medm}.
Auto-regressive video diffusion has emerged as a promising paradigm for streaming generation. CausVid~\cite{yin2025slow} distills a bidirectional video diffusion transformer into a causal auto-regressive generator with 4-step inference. Self Forcing~\cite{huang2025self} addresses exposure bias in auto-regressive video diffusion by conditioning on self-generated outputs during training. Diffusion Forcing~\cite{chen2024diffusion} unifies next-token prediction with full-sequence diffusion through independent per-token noise levels. Other auto-regressive approaches~\cite{sun2025ardiffusion,xie2024progressive,liu2024mardini,liu2025infvsr} and causal architectures~\cite{gao2024ca2,kim2024fifo} further demonstrate the potential of streaming diffusion.
Acceleration techniques include consistency models~\cite{luo2023latent,geng2025ect,song2023improved,kim2023consistency,heek2024multistep,lu2024simplifying}, advanced solvers~\cite{lu2022dpm,lu2022dpmsolverplusplus,zheng2023dpmsolver3}, flow-based methods~\cite{liu2024instaflow,jin2025pyramidal}, adversarial distillation~\cite{sauer2024adversarial,sauer2024fast,lin2024sdxl,xu2024ufogen}, distribution matching distillation~\cite{yin2024one,yin2024improved,zhou2024score}, other distillation approaches~\cite{salimans2022progressive,meng2023distillation,zhou2024simple,xie2024distillation,zhuang2025flashvsr,ren2024hyper,chen2025sana,zheng2024trajectory}, video-specific distillation~\cite{zhai2024motion}, and efficient architectures~\cite{huang2025instantvir}. Theoretical advances~\cite{wang2024phased,gu2025rectified} and recent image/offline distillation methods~\cite{sun2024pisasr,zhang2024degradationguidedonestepimagesuperresolution,wu2024one,wu2024seesr} exist. In contrast, \emph{Stream-DiffVSR} combines diffusion distillation with strictly online (past-only) causal temporal modeling to enable low-latency VSR.

\vspace{-.5em}
\section{Method}
\label{sec:method}
\vspace{-.5em}

We propose Stream-DiffVSR, a streamable auto-regressive diffusion framework for efficient video super-resolution (VSR). Its core innovation lies in an auto-regressive formulation that improves both temporal consistency and inference speed. The framework comprises: (1) a distilled few-step U-Net for accelerated diffusion inference, (2) Auto-regressive Temporal Guidance that conditions latent denoising on previously warped high-quality frames, (3) an Auto-regressive Temporal-aware Decoder that explicitly incorporates temporal cues, and (4) a staged training scheme with rollout distillation over full inference trajectories. Together, these components enable Stream-DiffVSR to produce stable and perceptually coherent videos.

\vspace{-.5em}
\subsection{Diffusion Models Preliminaries}
\vspace{-.5em}
Diffusion Models~\cite{ho2020denoising} transform complex data distributions into simpler Gaussian distributions via a forward diffusion process and reconstruct the original data using a learned reverse denoising process. The forward process gradually adds Gaussian noise to the initial data $x_0$, forming a Markov chain: \(q(x_t \mid x_{t-1}) = \mathcal{N}\!\bigl(x_t; \sqrt{1 - \beta_t}\, x_{t-1},\, \beta_t I\bigr)\) for \(t = 1, \dots, T\), where $\beta_t$ denotes a predefined noise schedule. At timestep $t$, the noised data $x_t$ can be directly sampled from the clean data $x_0$ as: \(x_t = \sqrt{\alpha_t}\, x_0 + \sqrt{1 - \alpha_t}\, \epsilon\), where \(\epsilon \sim \mathcal{N}(0, I)\) and \(\alpha_t = \prod_{i=1}^{t}(1 - \beta_i)\), where $\alpha_t = \prod_{i=1}^{t}(1 - \beta_i)$. The reverse process progressively removes noise from \(x_T\), reconstructing the original data \(x_0\) through a learned denoising operation modeled as a Markov chain, i.e., 
\(p_\theta(x_0, \dots, x_{T-1}\mid x_T) = \prod_{t=1}^{T} p_\theta(x_{t-1}\mid x_t)\). 
Each individual step is parameterized by a neural network-based denoising function 
\(p_\theta(x_{t-1}\mid x_t) = \mathcal{N}\!\bigl(x_{t-1}; \mu_\theta(x_t, t), \Sigma_\theta(t) I\bigr)\). 
Typically, the network predicts the noise component \(\epsilon_\theta(x_t, t)\), from which the denoising mean is estimated as 
\(\mu_\theta(x_t, t) = \tfrac{1}{\sqrt{\alpha_t}}\Bigl(x_t - \tfrac{1 - \alpha_t}{\sqrt{1 - \alpha_t}}\, \epsilon_\theta(x_t, t)\Bigr)\). Latent Diffusion Models (LDMs)~\cite{rombach2022high} further reduce computational complexity by projecting data into a lower-dimensional latent space using Variational Autoencoders (VAEs), significantly accelerating inference without sacrificing generative quality.

\vspace{-.5em}
\subsection{U-Net Rollout Distillation}
\vspace{-.5em}
We distill a pre-trained Stable Diffusion (SD) $\times$4 Upscaler~\cite{rombach2022high,Rombach_2022_CVPR}, originally designed for 50-step inference, into a 4-step variant that balances speed and perceptual quality. To mitigate the training–inference gap of timestep-sampling distillation, we adopt rollout distillation, where the U-Net performs the full 4-step denoising each iteration to obtain a clean latent. Detailed algorithms and implementation are provided in the supplementary material due to page limits.



Unlike conventional distillation that supervises random intermediate timesteps, our method applies loss only on the final denoised latent, ensuring the training trajectory mirrors inference and improving stability and alignment.


Our distillation requires no architectural changes. We train the U-Net by optimizing latent reconstruction with a loss that balances spatial accuracy, perceptual fidelity, and realism:
\begin{equation} \small
\begin{aligned}
\mathcal{L}_{\text{distill}} = &\, \| \mathbf{z}_{\text{den}} - \mathbf{z}_{\text{gt}} \|_2^2 \\
&+ \lambda_{\text{LPIPS}} \cdot \text{LPIPS}\left( D(\mathbf{z}_{\text{den}}), \mathbf{x}_{\text{gt}} \right) \\
&+ \lambda_{\text{GAN}} \cdot \mathcal{L}_{\text{GAN}}\left( D(\mathbf{z}_{\text{den}}) \right),
\end{aligned}
\label{unet_distill_math}
\end{equation}
where \(\mathbf{z}_{\text{den}}\) and \(\mathbf{z}_{\text{gt}}\) are the denoised and ground-truth latent representations. The decoder \(D(\cdot)\) maps latent features back to RGB space for perceptual (LPIPS) and adversarial (GAN) loss calculations, encouraging visually realistic outputs.



\begin{figure}[t]
    \centering
    \includegraphics[width=\columnwidth,height=0.5\columnwidth,keepaspectratio]{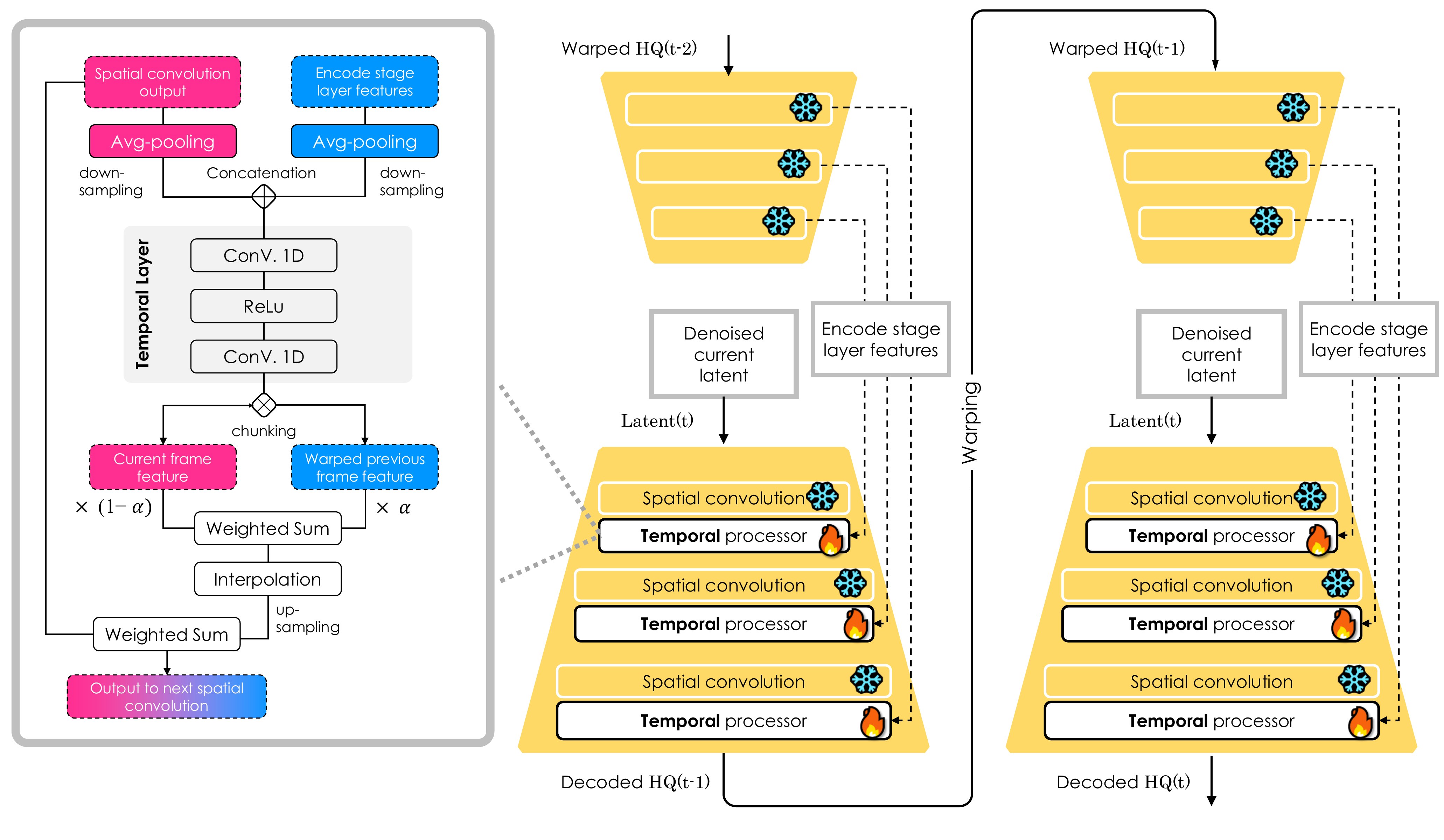}
\vspace{-4mm}
    \caption{
    \textbf{Overview of Auto-regressive Temporal-aware Decoder.} 
    Given the denoised latent and warped previous frame, our decoder enhances temporal consistency using temporal processor modules. This module aligns and fuses these features via interpolation, convolution, and weighted fusion, effectively stabilizing detail reconstruction when decoding into the final RGB frame.
    }
    \label{fig:autoencoder}
\end{figure}

\vspace{-.5em}
\subsection{Auto-regressive Temporal Guidance}
\vspace{-.5em}
Leveraging temporal information is essential for capturing dynamics and ensuring frame continuity in video super-resolution. However, extensive temporal reasoning often incurs significant computational overhead, increasing per-frame inference time and system latency. Thus, efficient online VSR requires carefully balancing temporal utilization and computational cost to support low-latency processing.


To this end, we propose \textbf{Auto-regressive Temporal Guidance (ARTG)}, which enforces temporal coherence during latent denoising. At each timestep~\(t\), the U-Net takes both the current noised latent \(z_t\) and the warped RGB frame from the previous output, \(\hat{x}_{t-1}^{\text{warp}} = \text{Warp}(x_{t-1}^{\text{SR}}, f_{t \leftarrow t-1})\), where \(f_{t \leftarrow t-1}\) is the optical flow from frame \(t{-}1\) to \(t\). The denoising prediction is then formulated as:
\vspace{-3pt}
\begin{equation} \small
\hat{\epsilon}_\theta = \text{UNet}(z_t, t, \hat{x}_{t-1}^{\text{warp}}),
\end{equation}
where the warped image $\hat{x}_{t-1}^{\text{warp}}$ serves as temporal conditioning input to guide the denoising process.

We train the \textbf{ARTG} module independently using consecutive pairs of low-quality and high-quality frames. The denoising U-Net and decoder are kept fixed during this stage, and the training objective focuses on reconstructing the target latent representation while preserving perceptual quality and visual realism. The total loss function is defined as:
\begin{equation} \label{artg_training_math} \small
\begin{aligned}
\mathcal{L}_{\text{ARTG}} = &\, \| \mathbf{z}_{\text{den}} - \mathbf{z}_{\text{gt}} \|_2^2 \\
&+ \lambda_{\text{LPIPS}} \cdot \text{LPIPS}(D(\mathbf{z}_{\text{den}}), \mathbf{x}_{\text{gt}}) \\
&+ \lambda_{\text{GAN}} \cdot \mathcal{L}_{\text{GAN}}(D(\mathbf{z}_{\text{den}})),
\end{aligned}
\end{equation}
where $\mathbf{z}_{\text{den}}$ denotes the denoised latent from DDIM updates with predicted noise $\hat{\boldsymbol{\epsilon}}_\theta$, and $\mathbf{z}_{\text{gt}}$ is the ground-truth latent. The decoder $D(\cdot)$ maps latents to RGB, producing $D(\mathbf{z}_{\text{den}})$ for comparison with the ground-truth image $\mathbf{x}_{\text{gt}}$. The latent $\ell_2$ loss enforces alignment, the perceptual loss preserves visual fidelity, and the adversarial loss promotes realism. This design leverages only past frames to propagate temporal context, improving consistency without additional latency.

\begin{figure*}[t]
    \centering
    \includegraphics[width=\linewidth]{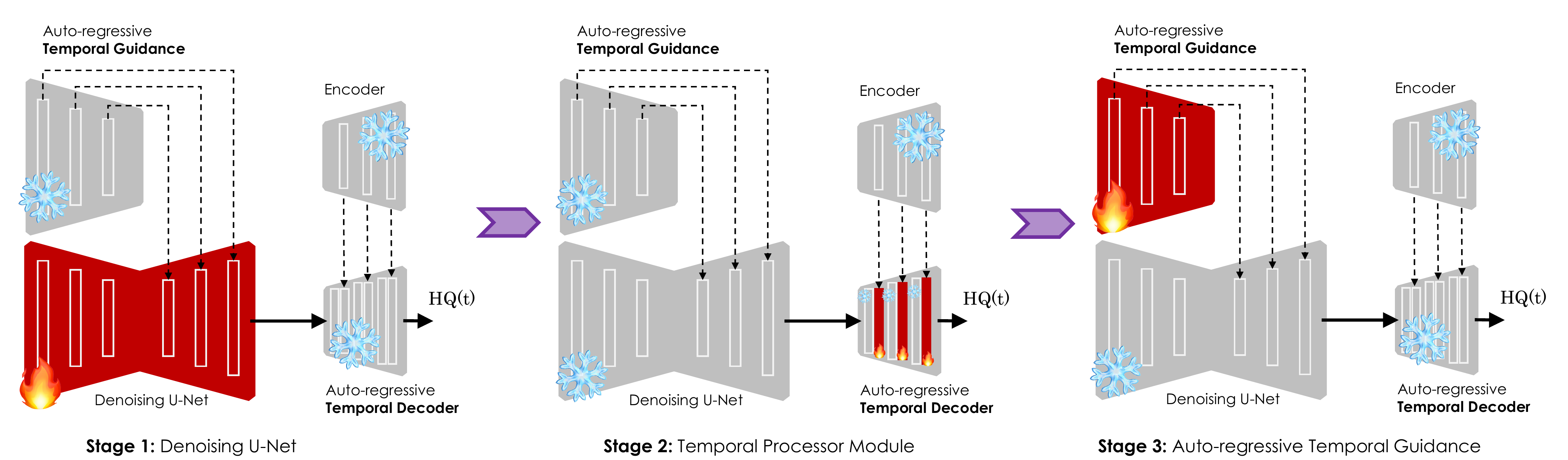}
    \vspace{-6mm}
    \caption{\textbf{Training pipeline of Stream-DiffVSR.} The training process consists of three sequential stages: (1) Distilling the denoising U-Net to reduce diffusion steps while maintaining perceptual quality with training objective (\ref{unet_distill_math}); (2) Training the Temporal Processor Module (TPM) within the decoder to enhance temporal consistency at the RGB level with training objective (\ref{artg_training_math}); (3) Training the Auto-Regressive Temporal Guidance (ARTG) module to leverage previously restored high-quality frames for improved temporal coherence with training objective (\ref{tpm_training_math}). Each module is trained separately before integrating them into the final framework.}
    \label{fig:training_stage}
\end{figure*}

\begin{figure*}[t]
    \centering
    \includegraphics[width=\linewidth]{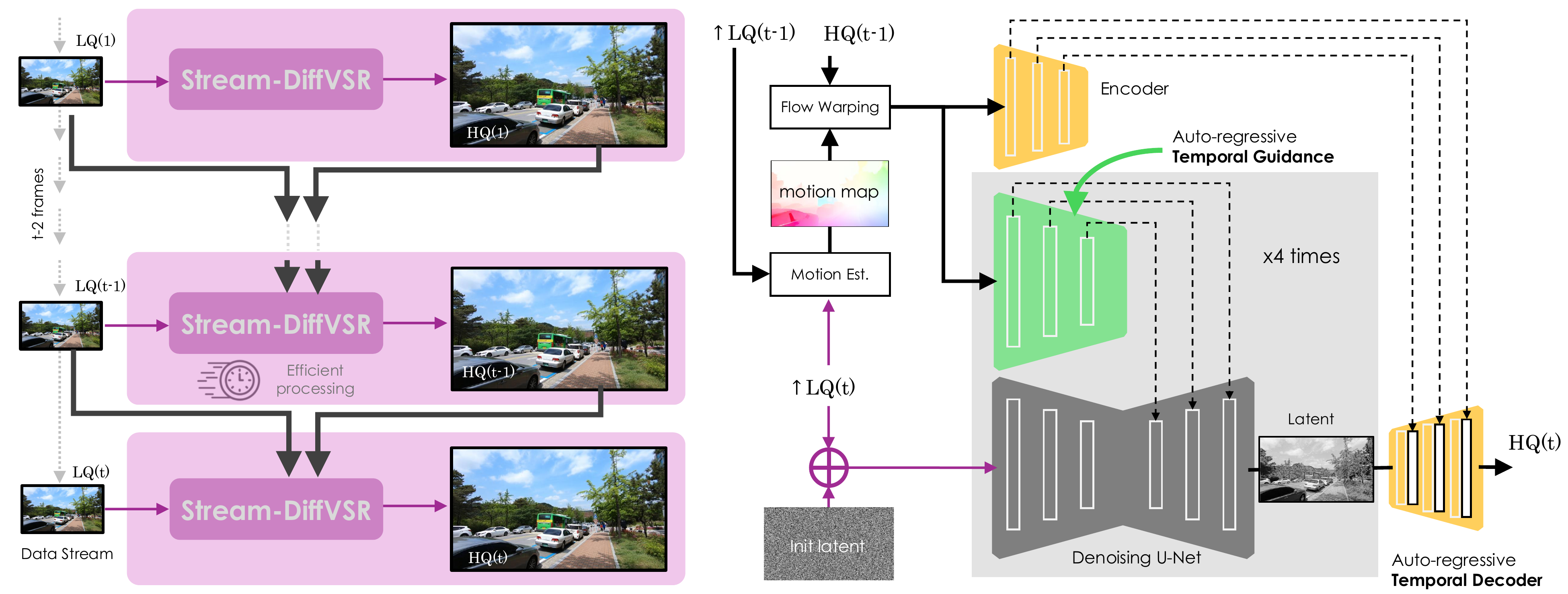}
\vspace{-6mm}
    \caption{\textbf{Overview of our pipeline.} Given a low-quality (LQ) input frame, we first initialize its latent representation and employ an autoregressive diffusion model composed of a distilled denoising U-Net, autoregressive temporal Guidance, and an autoregressive temporal Decoder. Temporal guidance utilizes flow-warped high-quality (HQ) results from the previous frame to condition the current frame’s latent denoising and decoding processes, significantly improving perceptual quality and temporal consistency in an efficient, online manner.}
    \label{fig:inference_pipeline}
\end{figure*}

\vspace{-.5em}
\subsection{Auto-regressive Temporal-aware Decoder} \label{sec:autoregressive}
\vspace{-.5em}
Although the \textbf{Auto-regressive Temporal Guidance (ARTG)} improves temporal consistency in the latent space, the features produced by the Stable Diffusion $\times$4 Upscaler remain at one-quarter of the target resolution. This mismatch may introduce decoding artifacts or misalignment in dynamic scenes.

To address this issue, we propose an \textbf{Auto-regressive Temporal-aware Decoder} that incorporates temporal context into decoding to enhance spatial fidelity and temporal consistency. At timestep~\(t\), the decoder takes the denoised latent \( \mathbf{z}_t^{\text{den}} \) and the aligned feature \( \hat{\mathbf{f}}_{t-1} \) derived from the previous super-resolved frame. Specifically, we compute:
\begin{equation} \small
\hat{\mathbf{x}}_{t-1}^{\text{warp}} = \text{Warp}(\mathbf{x}_{t-1}^{\text{SR}}, f_{t \leftarrow t-1}),\,\,\,\,
\hat{\mathbf{f}}_{t-1} = \text{Enc}(\hat{\mathbf{x}}_{t-1}^{\text{warp}}),
\end{equation}
where \( \mathbf{x}_{t-1}^{\text{SR}} \) is the previously generated RGB output, \( f_{t \leftarrow t-1} \) is the optical flow from frame \( t-1 \) to \( t \), and \( \text{Enc}(\cdot) \) is a frozen encoder that projects the warped image into the latent feature space.

The decoder then synthesizes the current frame using:
\begin{equation} \small
\mathbf{x}_t^{\text{SR}} = \text{Decoder}(\mathbf{z}_t^{\text{den}}, \hat{\mathbf{f}}_{t-1}).
\end{equation}

\vspace{-1mm}
We adopt a multi-scale fusion strategy inside the decoder to combine current spatial information and prior temporal features across multiple resolution levels, as illustrated in Fig.~\ref{fig:autoencoder}. This design helps reinforce temporal coherence while recovering fine spatial details.


\vspace{-.5em}
\paragraph{Temporal Processor Module (TPM).}
We integrate TPM after each spatial convolutional layer in the decoder to explicitly inject temporal coherence, enhancing stability and continuity of reconstructed frames. These modules utilize latent features from the current frame and warped features from the previous frame, optimizing temporal consistency independently from spatial reconstruction. Our training objective for the TPM is defined as:
\begin{equation} \label{tpm_training_math} \small
\begin{aligned}
\mathcal{L}_{\text{TPM}} =& \; \mathcal{L}_{\text{rec}}(\mathbf{x}_t^{\text{rec}}, \mathbf{x}_t^{\text{GT}}) \\
&+  \lambda_{\text{flow}}  \left\| \text{OF}(\mathbf{x}_t^{\text{rec}}, \mathbf{x}_{t-1}^{\text{rec}}) - \text{OF}(\mathbf{x}_t^{\text{GT}}, \mathbf{x}_{t-1}^{\text{GT}}) \right\|_2^2 \\
&+ \lambda_{\text{GAN}} \mathcal{L}_{\text{GAN}}(\mathbf{x}_t^{\text{rec}}) \\
&+ \lambda_{\text{LPIPS}} \text{LPIPS}(\mathbf{x}_t^{\text{rec}}, \mathbf{x}_t^{\text{GT}}),
\end{aligned}
\end{equation}
where \( \mathbf{x}_t^{\text{SR}} \in \mathbb{R}^{3 \times H \times W} \) is the predicted frame at time~\(t\), and \( \mathbf{x}_t^{\text{GT}} \) is the ground-truth frame. The reconstruction loss \(\mathcal{L}_{\text{rec}} = \text{SmoothL1}(\mathbf{x}_t^{\text{rec}}, \mathbf{x}_t^{\text{GT}})\) enforces spatial fidelity, the adversarial loss \(\mathcal{L}_{\text{GAN}}\) improves realism, and the optical-flow term \(\text{OF}(\cdot,\cdot)\) reduces temporal discrepancies, yielding consistent and perceptually faithful outputs.



\vspace{-.5em}
\subsection{Training and Inference Stages} \label{sec:training_inference_stages}
\vspace{-.5em}
Our training pipeline consists of three independent stages (\cref{fig:training_stage}), while our inference process and the Auto-Regressive Diffusion-based VSR algorithm are illustrated in \cref{fig:inference_pipeline} and detailed in the appendix due to page constraints, respectively.


\vspace{-.5em}
\paragraph{Distilling the Denoising U-Net.}
We first distill the denoising U-Net using pairs of low-quality (LQ) and high-quality (HQ) frames to optimize per-frame super-resolution and latent-space consistency.

\vspace{-.5em}
\paragraph{Training the Temporal Processor Module (TPM).}
In parallel, we train the Temporal Processor Module (TPM) in the decoder using ground-truth frames, keeping all other weights fixed. This enhances the decoder’s capability to incorporate temporal information into the final RGB reconstruction.

\vspace{-.5em}
\paragraph{Training Auto-regressive Temporal Guidance.}
After training and freezing the U-Net and decoder, we train the ARTG, which leverages flow-aligned previous outputs to enhance temporal coherence without degrading spatial quality. This staged training strategy progressively refines spatial fidelity, latent consistency, and temporal smoothness in a decoupled manner.
%

\vspace{-.5em}
\paragraph{Inference.} 
Given a sequence of low-quality (LQ) frames, our method auto-regressively generates high-quality (HQ) outputs. For each frame $t$, denoising is conditioned on the previous output $HQ_{t-1}$, warped via optical flow to capture temporal motion. To balance quality and efficiency, we employ a \textbf{4-step DDIM} scheme using a distilled U-Net. By combining motion alignment with reduced denoising steps, our inference pipeline achieves efficient and stable temporal consistency.

\vspace{-.5em}
\section{Experiment}
\label{sec:experiment}
\vspace{-.5em}

Due to space limitations, we provide the experimental setup in the appendix.


\begin{table*}[t]
\caption{
\textbf{Quantitative comparison on the REDS4 dataset.} 
We compare bidirectional/offline and unidirectional/online methods. \(\uparrow\)/\(\downarrow\) indicate higher/lower is better. \textbf{Dir.}: \textbf{B} for bidirectional/offline, \textbf{U} for unidirectional/online. \textbf{tLP} and \textbf{tOF} are scaled by 100× and 10×. Best and second-best results within each direction group are marked in \textcolor{red}{red} and \textcolor{blue}{blue}. \textcolor{gray}{Gray-shaded entries are reported from FlashVSR}.
}

\label{tab:reds4_all_methods_performance}
\vspace{-3mm}
\centering
\small
\setlength{\tabcolsep}{3pt}
\resizebox{\textwidth}{!}{%
\begin{tabular}{llccccccccccc}
\toprule
Dir. & Method & PSNR$\uparrow$ & SSIM$\uparrow$ & LPIPS$\downarrow$ & DISTS$\downarrow$ & MUSIQ$\uparrow$ & NIQE$\downarrow$ & NRQM$\uparrow$ & BRISQUE$\downarrow$ & VMAF$\uparrow$ & tLP$\downarrow$ & tOF$\downarrow$ \\
\midrule
\multicolumn{13}{c}{\textbf{Bidirectional/Offline Methods}} \\
\midrule
-  & Bicubic & 25.501 & 0.712 & 0.460 & 0.187 & 27.362 & 7.360 & 3.459 & 60.256 & 46.132 & 21.603 & 4.241 \\
B  & BasicVSR++ & 32.386 & 0.907 & 0.132 & 0.069 & 67.013 & 3.850 & 6.363 & 38.641 & 99.580 & 9.017 & 2.490 \\
B  & RealBasicVSR & 27.042 & 0.778 & 0.134 & 0.065 & 67.033 & \textcolor{red}{2.530} & \textcolor{blue}{6.769} & 18.046 & 88.197 & \textcolor{blue}{6.422} & 4.759 \\
B  & RVRT & \textcolor{blue}{32.701} & \textcolor{blue}{0.911} & 0.130 & 0.067 & \textcolor{blue}{67.251} & 3.793 & 6.366 & 38.038 & \textcolor{red}{99.681} & 9.133 & \textcolor{blue}{2.421} \\
B  & MIA-VSR & \textcolor{red}{32.790} & \textcolor{red}{0.912} & \textcolor{blue}{0.123} & \textcolor{blue}{0.064} & \textcolor{red}{68.140} & 3.742 & 6.451 & 37.099 & \textcolor{blue}{99.669} & 8.870 & \textcolor{red}{2.354} \\
B  & StableVSR & 27.928 & 0.793 & \textcolor{red}{0.102} & \textcolor{red}{0.047} & 67.058 & \textcolor{blue}{2.713} & \textcolor{red}{6.960} & \textcolor{blue}{16.249} & 95.303 & \textcolor{red}{5.755} & 2.742 \\
B  & MGLD-VSR & 26.53 & 0.749 & 0.151 & 0.065 & 66.081 & 2.972 & 6.701 & \textcolor{red}{15.291} & 78.255 & 18.139 & 5.910 \\
B  & UAV & 23.077 & 0.608 & 0.495 & 0.249 & 30.912 & 5.917 & 3.666 & 39.751 & 20.600 & 30.546 & 13.582 \\
\rowcolor{black!10}
B  & UAV & 24.840 & 0.644 & 0.41 & -- & 53.000 & 3.104 & -- & -- & -- & -- & -- \\
B  & SeedVR2 & 22.774 & 0.661 & 0.246 & 0.111 & 65.011 & 2.812 & 6.707 & 29.948 & 64.447 & 30.296 & 12.79 \\
\rowcolor{black!10}
B  & SeedVR2 & 24.830 & 0.704 & 0.302 & -- & 61.830 & 3.066 & -- & -- & -- & -- & -- \\
B  & DOVE & 25.027 & 0.692 & 0.291 & 0.149 & 54.060 & 4.087 & 5.753 & 25.606 & 58.329 & 10.260 & 10.960 \\
B  & VEnhancer & 20.712 & 0.560 & 0.384 & 0.156 & 56.932 & 4.142 & 5.567 & 24.906 & 24.743 & 32.913 & 16.643 \\
\midrule
\multicolumn{13}{c}{\textbf{Unidirectional/Online Methods}} \\
\midrule
U  & TMP & \textcolor{red}{30.672} & \textcolor{red}{0.871} & 0.194 & 0.090 & 63.818 & 4.378 & 5.796 & 43.394 & \textcolor{red}{98.586} & \textcolor{blue}{10.424} & \textcolor{red}{2.480} \\
U  & RealViformer & 26.763 & 0.761 & 0.129 & 0.065 & 64.585 & \textcolor{blue}{2.731} & \textcolor{blue}{7.028} & 17.272 & 60.509 & 11.261 & 4.037 \\
U  & StableVSR* & 27.174 & 0.763 & \textcolor{blue}{0.111} & \textcolor{red}{0.051} & \textcolor{red}{66.428} & \textcolor{red}{2.572} & 6.944 & \textcolor{red}{15.805} & 88.675 & 11.107 & 3.925 \\
U  & FlashVSR & 21.484 & 0.569 & 0.283 & 0.124 & 62.178 & 6.574 & 6.574 & 25.821 & 40.556 & 15.69 & 20.548 \\
\rowcolor{black!10}
U  & FlashVSR & 23.920 & 0.649 & 0.343 & -- & 68.97 & 2.425 & -- & -- & -- & -- & -- \\
U  & FlashVSR-tiny & 21.887 & 0.577 & 0.300 & 0.141 & 55.204 & 3.463 & 6.188 & 28.589 & 38.634 & 17.838 & 23.992 \\
\rowcolor{black!10}
U  & FlashVSR-tiny & 24.110 & 0.651 & 0.343 & -- & 67.43 & 2.680 & -- & -- & -- & -- & -- \\
U  & Ours & \textcolor{blue}{27.256} & \textcolor{blue}{0.766} & \textcolor{red}{0.099} & \textcolor{blue}{0.062} & \textcolor{blue}{65.586} & 3.114 & \textcolor{red}{7.055} & \textcolor{blue}{17.117} & \textcolor{blue}{88.751} & \textcolor{red}{4.265} & \textcolor{blue}{3.620} \\
\bottomrule
\end{tabular}%
}
\end{table*}

\begin{table*}[t]
\caption{\textbf{Efficiency comparison on the REDS4 dataset.} Runtime, TTFF, and peak memory are measured per 720p frame; memory-intensive methods are measured on an RTX Pro 6000, others on an RTX 4090. \textbf{TTFF} (time-to-first-frame) is the wait from the first input frame to the first output frame, measured over 100-frame sequences; for offline/bidirectional methods this scales with sequence length. Best and second-best results within each GPU/direction group are marked in \textcolor{red}{red} and \textcolor{blue}{blue}.}
\label{tab:reds4_all_methods_efficiency}
\vspace{-3mm}
\centering
\small
\setlength{\tabcolsep}{4pt}
\resizebox{\textwidth}{!}{%
\begin{tabular}{ll ccc @{\hspace{2.5em}} ll ccc}
\toprule

Dir. & Method & Runtime (s)$\downarrow$ & TTFF (s)$\downarrow$ & Mem (GB)$\downarrow$ &
Dir. & Method & Runtime (s)$\downarrow$ & TTFF (s)$\downarrow$ & Mem (GB)$\downarrow$ \\
\midrule
\multicolumn{10}{c}{\textbf{Measured on an NVIDIA RTX 4090 GPU}} \\
\midrule
B & BasicVSR++ & \textcolor{blue}{0.098} & 9.8 & - & U & TMP & \textcolor{red}{0.041}  & \textcolor{red}{0.041} & - \\
B & RealBasicVSR & \textcolor{red}{0.064} & \textcolor{blue}{6.4} & - & U & RealViformer & \textcolor{blue}{0.099} & 9.9 & - \\
B & RVRT & 0.498 & \textcolor{red}{2.49} & - & U & StableVSR* & 46.2 & 4620 & - \\
B & MIA-VSR & 0.768 & 76.8 & - & U & Ours & 0.328 & \textcolor{blue}{0.328} & - \\
B & StableVSR & 46.2 & 4620 & - & & & & & \\
B & MGLD-VSR & 43.6 & 218 & - & & & & & \\
\midrule
\multicolumn{10}{c}{\textbf{Measured on an NVIDIA RTX Pro 6000 GPU}} \\
\midrule
B & UAV & 6.017 & 601.740 & 57.094 & U & FlashVSR & 0.283 & 1.698 & 19.630 \\
B & SeedVR2  & \textcolor{red}{0.479} & \textcolor{red}{47.840} & 69.592 & U & FlashVSR-tiny  & \textcolor{red}{0.157} & \textcolor{blue}{0.942} & \textcolor{red}{12.390} \\
B & DOVE  & \textcolor{blue}{0.760} & \textcolor{blue}{75.983}  & \textcolor{red}{42.208} & U & Ours  & \textcolor{blue}{0.243} & \textcolor{red}{0.243}  & \textcolor{blue}{19.580} \\
B & VEnhancer   & 0.672 & 671.650 & \textcolor{blue}{53.200} & & & & & \\
\bottomrule
\end{tabular}%
}
\end{table*}

\begin{table*}[t]
\caption{
\textbf{Quantitative comparison on the Vimeo-90K-T dataset.}}
\label{tab:vimeo_all_methods_performance}
\vspace{-3mm}
\centering
\small
\setlength{\tabcolsep}{3pt}
\resizebox{\textwidth}{!}{%
\begin{tabular}{llccccccccccc}
\toprule
Dir. & Method & PSNR$\uparrow$ & SSIM$\uparrow$ & LPIPS$\downarrow$ & DISTS$\downarrow$ & MUSIQ$\uparrow$ & NIQE$\downarrow$ & NRQM$\uparrow$ & BRISQUE$\downarrow$ & VMAF$\uparrow$ & tLP$\downarrow$ & tOF$\downarrow$ \\
\midrule
\multicolumn{13}{c}{\textbf{Bidirectional/Offline Methods}} \\
\midrule
- & Bicubic & 29.282 & 0.864 & 0.297 & 0.209 & 23.433 & 8.735 & 3.588 & 61.714 & 42.928 & 11.606 & 2.49 \\
B & BasicVSR++ & 37.479 & 0.956 & 0.098 & 0.117 & 51.940 & 7.077 & 5.509 & 47.792 & 92.905 & \textcolor{red}{4.691} & 1.57 \\
B & RealBasicVSR & 29.388 & 0.857 & 0.156 & 0.149 & 56.986 & 5.069 & 7.413 & 23.822 & 79.781 & 10.947 & 3.46 \\
B & RVRT & \textcolor{red}{37.815} & 0.955 & \textcolor{blue}{0.093} & \textcolor{blue}{0.105} & 49.937 & 7.205 & 5.393 & 48.352 & \textcolor{blue}{94.660} & 4.873 & \textcolor{blue}{1.429} \\
B & MIA-VSR & \textcolor{blue}{37.598} & \textcolor{red}{0.957} & \textcolor{red}{0.086} & \textcolor{red}{0.101} & 51.402 & 7.116 & 5.569 & 47.865 & \textcolor{red}{95.113} & \textcolor{blue}{4.696} & \textcolor{red}{1.419} \\
B & StableVSR & 31.823 & 0.878 & 0.095 & 0.111 & 54.582 & \textcolor{blue}{4.745} & 7.265 & \textcolor{red}{20.039} & 86.936 & 26.224 & 3.108 \\
B & MGLD-VSR & 29.651 & 0.865 & 0.151 & 0.137 & 57.788 & 5.340 & 7.217 & 20.761 & 71.509 & 12.550 & 4.661 \\
B & UAV & 25.263 & 0.72 & 0.226 & 0.177 & 55.562 & 5.602 & 7.247 & 24.766 & 46.489 & 9.546 & 7.576 \\
B & SeedVR2 & 24.585 & 0.719 & 0.177 & 0.152 & \textcolor{red}{58.809} & \textcolor{red}{4.204} & \textcolor{red}{8.107} & \textcolor{blue}{20.447} & 41.430 & 21.191 & 12.460 \\
B & DOVE & 28.611 & 0.834 & 0.12 & 0.116 & 57.749 & 5.734 & 7.182 & 23.388 & 77.997 & 8.214 & 4.387 \\
B & VEnhancer & 23.712 & 0.712 & 0.234 & 0.189 & 57.822 & 6.175 & \textcolor{blue}{7.910} & 34.906 & 60.920 & 32.913 & 16.643 \\
\midrule
\multicolumn{13}{c}{\textbf{Unidirectional/Online Methods}} \\
\midrule
U & TMP & \textcolor{red}{36.482} & \textcolor{red}{0.946} & 0.109 & 0.118 & 48.374 & 7.368 & 5.096 & 49.192 & \textcolor{red}{92.001} & \textcolor{blue}{4.870} & \textcolor{red}{1.603} \\
U & RealViformer & 30.291 & 0.877 & 0.130 & 0.140 & 53.107 & 5.515 & 6.711 & \textcolor{blue}{24.628} & 54.689 & 8.232 & 2.769 \\
U & StableVSR* & 31.729 & 0.875 & \textcolor{blue}{0.072} & \textcolor{blue}{0.113} & 54.447 & \textcolor{blue}{4.698} & 7.280 & \textcolor{red}{19.836} & 86.162 & 30.858 & 3.144 \\
U & FlashVSR & 26.114 & 0.789 & 0.136 & 0.118 & \textcolor{red}{59.451} & 5.249 & \textcolor{blue}{7.317} & 29.288 & 78.575 & 19.018 & 9.351 \\
U & FlashVSR-tiny & 26.154 & 0.791 & 0.136 & 0.121 & \textcolor{blue}{57.845} & 5.249 & \textcolor{blue}{7.317} & 29.288 & 76.376 & 19.881 & 9.793 \\
U & Ours & \textcolor{blue}{32.593} & \textcolor{blue}{0.900} & \textcolor{red}{0.056} & \textcolor{red}{0.105} & 52.755 & \textcolor{red}{4.403} & \textcolor{red}{7.672} & 29.297 & \textcolor{blue}{88.311} & \textcolor{red}{4.307} & \textcolor{blue}{2.689} \\
\bottomrule
\end{tabular}%
}
\end{table*}
\begin{table*}[t]
\caption{
\textbf{Efficiency comparison on the Vimeo-90K-T dataset.} Runtime, TTFF, and peak memory are measured per 448$\times$256 frame.}
\label{tab:vimeo_all_methods_efficiency}
\vspace{-3mm}
\centering
\small
\setlength{\tabcolsep}{4pt}
\resizebox{\textwidth}{!}{%
\begin{tabular}{ll ccc @{\hspace{2.5em}} ll ccc}
\toprule
Dir. & Method & Runtime (s)$\downarrow$ & TTFF (s)$\downarrow$ & Mem (GB)$\downarrow$ &
Dir. & Method & Runtime (s)$\downarrow$ & TTFF (s)$\downarrow$ & Mem (GB)$\downarrow$ \\
\midrule
\multicolumn{10}{c}{\textbf{Measured on an NVIDIA RTX 4090 GPU}} \\
\midrule
B & BasicVSR++ & \textcolor{blue}{0.012} & \textcolor{blue}{0.084} & - & U & TMP & \textcolor{red}{0.006}  & \textcolor{red}{0.006} & - \\
B & RealBasicVSR & \textcolor{red}{0.008}  & \textcolor{red}{0.056} & - & U & RealViformer & \textcolor{blue}{0.013} & 0.091 & - \\
B & RVRT & 0.061 & 0.305 & - & U & StableVSR* & 5.749 & 40.243 & - \\
B & MIA-VSR & 0.096 & 0.672 & - & U & Ours & 0.041 & \textcolor{blue}{0.041} & - \\
B & StableVSR & 5.749 & 40.243 & - & & & & & \\
B & MGLD-VSR & 5.426 & 27.130 & - & & & & & \\
\midrule
\multicolumn{10}{c}{\textbf{Measured on an NVIDIA RTX Pro 6000 GPU}} \\
\midrule
B & UAV & 0.499 & 3.490 & 37.987 & U & FlashVSR & \textcolor{blue}{0.025} & 0.150 & \textcolor{blue}{5.340}  \\
B & SeedVR2 & \textcolor{red}{0.039}  & \textcolor{red}{0.270} & 41.366 & U & FlashVSR-tiny   & \textcolor{red}{0.019} & \textcolor{blue}{0.114} & \textcolor{red}{4.280}   \\
B & DOVE & \textcolor{blue}{0.223} & \textcolor{blue}{1.560} & \textcolor{red}{31.934} & U & Ours & 0.036 & \textcolor{red}{0.036}  & 9.720 \\
B & VEnhancer & 1.834 & 12.840 & \textcolor{blue}{35.855} & & & & & \\
\bottomrule
\end{tabular}%
}
\end{table*}

We quantitatively compare Stream-DiffVSR with state-of-the-art VSR methods on REDS4~\cite{Nah_2019_CVPR_REDS}, Vimeo-90K-T~\cite{xue2019video} and VideoLQ~\cite{chan2022investigating}, covering diverse scene content and motion characteristics. \cref{tab:reds4_all_methods_performance,tab:vimeo_all_methods_performance,tab:videolq_all_methods_performance} report perceptual and temporal-consistency metrics across CNN-, Transformer-, and diffusion-based approaches under both bidirectional (offline) and unidirectional (online) settings, including comparisons against memory-intensive diffusion-based baselines. \cref{tab:reds4_all_methods_efficiency,tab:vimeo_all_methods_efficiency,tab:videolq_efficiency} report the corresponding runtime, Time-to-First-Frame, and peak memory for each method, further highlighting Stream-DiffVSR's quality--latency trade-off under practical memory budgets.

On REDS4, Stream-DiffVSR achieves strong perceptual quality (LPIPS=0.099) compared with representative CNN (BasicVSR++~\cite{chan2022basicvsr++}, RealBasicVSR~\cite{chan2022investigating}), Transformer (RVRT~\cite{liang2022recurrent}), and diffusion-based baselines (StableVSR~\cite{rota2024enhancing}, MGLD-VSR~\cite{yang2024motion}), while maintaining competitive temporal consistency (tLP=4.198, tOF=3.638). Importantly, these results are obtained with substantially lower runtime (0.328s/frame vs.\ 43--46s/frame for diffusion-based baselines), matching the low-latency objective.

On Vimeo-90K-T, Stream-DiffVSR similarly attains favorable perceptual performance (LPIPS=0.056, DISTS=0.105) and improved temporal consistency (tLP=4.307, tOF=2.689) with a runtime of 0.041s/frame, supporting efficient causal inference for online/streaming usage.

\subsection{Generalization to Mixed Degradations}
\label{sec:mixed_degradation}

The REDS4 and Vimeo-90K-T evaluations above use bicubic downsampling, which isolates throughput, latency, and temporal-consistency comparisons from confounds introduced by stochastic degradation pipelines. To examine whether this controlled setting limits Stream-DiffVSR to a narrow degradation regime, we additionally fine-tune our model on REDS using the degradation pipeline introduced by DOVE~\cite{zhang2025dove} to synthesize realistic low-quality/high-quality pairs covering blur, noise, and compression artifacts. \cref{tab:mixed_degradation_comparison} compares the fine-tuned model against real-world-oriented baselines under this setting. Stream-DiffVSR remains competitive in perceptual quality (PSNR, SSIM, LPIPS) and attains the best or second-best temporal consistency (tLP, tOF) among online  methods, indicating that our architecture and training procedure are not tied to the bicubic setting used elsewhere in this paper. We further evaluate this fine-tuned model on the real-world VideoLQ in \cref{tab:videolq_all_methods_performance,tab:videolq_efficiency}, providing additional evidence of generalization in the absence of paired ground truth.

\begin{table}[h]
\caption{\textbf{REDS4 comparison under multi-degradations against real-world-oriented baselines.}}
\label{tab:mixed_degradation_comparison}
\vspace{-3mm}
\centering
\scriptsize
\setlength{\tabcolsep}{4pt}
\resizebox{\columnwidth}{!}{%
\begin{tabular}{llcccccccc}
\toprule
Dir. & Method & PSNR$\uparrow$ & SSIM$\uparrow$ & LPIPS$\downarrow$ & DISTS$\downarrow$ & MUSIQ$\uparrow$ & tLP$\downarrow$ & tOF$\downarrow$ \\
\midrule
B  & MGLD-VSR & 23.305 & 0.594 & 0.246 & 0.112 & 61.139 & 37.737 & 39.039 \\
B  & UAV & 21.721 & 0.534 & 0.419 & 0.22 & 45.714 & 14.687 & 30.589 \\
B  & SeedVR2 & 22.043 & 0.587 & 0.28 & 0.111 & 62.213 & 30.273 & 14.906 \\
B  & DOVE & 23.128 & 0.608 & 0.383 & 0.207 & 52.498 & 11.129 & 16.869 \\
\midrule
U  & FlashVSR & 21.163 & 0.527 & \textcolor{blue}{0.287} & \textcolor{red}{0.124} & \textcolor{red}{63.891} & \textcolor{red}{14.723} & \textcolor{blue}{24.353} \\
U  & FlashVSR-tiny & \textcolor{blue}{21.462} & \textcolor{blue}{0.536} & 0.307 & 0.142 & 57.593 & 20.221 & 27.106\\
U  & Ours & \textcolor{red}{22.377} & \textcolor{red}{0.549} & \textcolor{red}{0.22} & \textcolor{blue}{0.128} & \textcolor{blue}{63.222} & \textcolor{blue}{14.797} & \textcolor{red}{16.785}\\
\bottomrule
\end{tabular}%
}
\end{table}
\begin{table*}[!htb]
\centering
\caption{\textbf{Quantitative comparison on the VideoLQ dataset.} VideoLQ is a real-world benchmark without ground-truth references; we therefore report no-reference quality metrics only. Best and second-best results within each direction group are marked in red and blue. \textcolor{gray}{Gray text indicates entries reported from FlashVSR}.}
\label{tab:videolq_all_methods_performance}
\vspace{-3mm}
\small
\setlength{\tabcolsep}{4pt}
\resizebox{\textwidth}{!}{%
\begin{tabular}{ll ccc @{\hspace{2.5em}} ll ccc}
\toprule
Dir. & Method & NIQE$\downarrow$ & NRQM$\uparrow$ & BRISQUE$\downarrow$ &
Dir. & Method & NIQE$\downarrow$ & NRQM$\uparrow$ & BRISQUE$\downarrow$ \\
\midrule
B & BasicVSR++    & 5.909                   & 3.745                   & 56.800 & U & TMP            & 6.751                   & 3.511                   & 59.841 \\
B & RealBasicVSR  & \textcolor{red}{3.973}  & 6.095                   & 30.158 & U & RealViformer   & 4.070                   & 6.066                   & 28.266 \\
B & RVRT          & 6.939                   & 3.493                   & 60.557 & U & StableVSR*     & \textcolor{blue}{3.982} & \textcolor{blue}{6.122} & \textcolor{blue}{23.814} \\
B & MIA-VSR       & 5.860                   & 3.810                   & 58.513 & U & FlashVSR       & --                      & --                      & -- \\
B & StableVSR     & \textcolor{red}{3.973}  & \textcolor{red}{6.154}  & \textcolor{red}{22.973} & \textcolor{gray}{U} & \textcolor{gray}{FlashVSR} & \textcolor{gray}{3.803} & \textcolor{gray}{--} & \textcolor{gray}{--} \\
B & MGLD-VSR      & 4.163                   & 5.761                   & \textcolor{blue}{29.497} & U & FlashVSR-tiny  & 4.569                   & 5.164                   & 42.514 \\
B & UAV           & 6.299                   & 3.652                   & 44.139 & \textcolor{gray}{U} & \textcolor{gray}{FlashVSR-tiny} & \textcolor{gray}{4.070} & \textcolor{gray}{--} & \textcolor{gray}{--} \\
\textcolor{gray}{B} & \textcolor{gray}{UAV} & \textcolor{gray}{4.889} & \textcolor{gray}{--} & \textcolor{gray}{--} & U & Ours           & \textcolor{red}{3.929}  & \textcolor{red}{6.140}  & \textcolor{red}{23.176} \\
B & SeedVR2       & \textcolor{blue}{4.661} & \textcolor{blue}{5.523} & 37.975 &   &                &                         &                         &        \\
\textcolor{gray}{B} & \textcolor{gray}{SeedVR2} & \textcolor{gray}{5.205} & \textcolor{gray}{--} & \textcolor{gray}{--} &   &                &                         &                         &        \\
B & DOVE          & 5.090                   & 5.214                   & 36.631 &   &                &                         &                         &        \\
B & VEnhancer     & 6.221                   & 3.85                    & 48.1   &   &                &                         &                         &        \\
\bottomrule
\end{tabular}%
\vspace{-4mm}
}
\end{table*}
\begin{table*}[!htb]
\centering
\caption{\textbf{Efficiency comparison on the VideoLQ dataset under a single NVIDIA RTX Pro 6000 GPU.}}
\label{tab:videolq_efficiency}
\vspace{-3mm}
\small
\setlength{\tabcolsep}{4pt}
\resizebox{\textwidth}{!}{%
\begin{tabular}{ll ccc @{\hspace{2.5em}} ll ccc}
\toprule
Dir. & Method & Runtime (s)$\downarrow$ & TTFF (s)$\downarrow$ & Mem (GB)$\downarrow$ &
Dir. & Method & Runtime (s)$\downarrow$ & TTFF (s)$\downarrow$ & Mem (GB)$\downarrow$ \\
\midrule
B & VEnhancer & 9.544 & 477.237 & \textcolor{blue}{47.985} & U & FlashVSR      & --                      & --                      & OOM \\
B & SeedVR2   & \textcolor{red}{1.126} & \textcolor{red}{56.28} & 76.094 & U & FlashVSR-tiny & \textcolor{red}{0.204}  & \textcolor{blue}{1.224} & \textcolor{blue}{44.180} \\
B & UAV       & 8.081 & 404.07 & 55.897 & U & Ours          & \textcolor{blue}{0.454} & \textcolor{red}{0.454}  & \textcolor{red}{22.800} \\
B & DOVE      & \textcolor{blue}{1.735} & \textcolor{blue}{86.774} & \textcolor{red}{46.344} & & & & & \\
\bottomrule
\end{tabular}%
}
\end{table*}

In addition to speed, Stream-DiffVSR achieves a markedly lower memory footprint. We emphasize that Stream-DiffVSR performs strictly frame-by-frame causal inference, whereas FlashVSR~\cite{zhuang2025flashvsr} adopts chunk-wise streaming. Consequently, FlashVSR’s per-frame runtime can appear low, but its time-to-first-frame is bounded by chunk buffering, while Stream-DiffVSR avoids sequence-level waiting and provides immediate frame-level response. As shown in \cref{tab:reds4_all_methods_efficiency,tab:vimeo_all_methods_efficiency,tab:videolq_efficiency}, prior diffusion-based VSR methods such as DOVE~\cite{zhang2025dove}, SeedVR2~\cite{wang2025seedvr2}, and Upscale-A-Video(UAV)~\cite{zhou2024upscale} often require large GPU memory budgets (e.g., exceeding 42\,GB on REDS4/VideoLQ under our settings) or run into OOM on single RTX 4090. In contrast, Stream-DiffVSR operates within 19.58\,GB and 22.80\,GB peak memory on REDS4 and VideoLQ, respectively, and consistently achieves lower runtime than every memory-intensive baseline across both datasets, with speedups ranging from approximately $2\times$ over the closest competitor (SeedVR2) to over $20\times$ over the slowest baselines (UAV, VEnhancer), underscoring its efficiency and deployability.

Results on VideoLQ further show consistent perceptual and temporal performance, indicating stable behavior across diverse content and motion patterns.

\begin{figure*}[t]
    \centering
    \includegraphics[width=\linewidth]{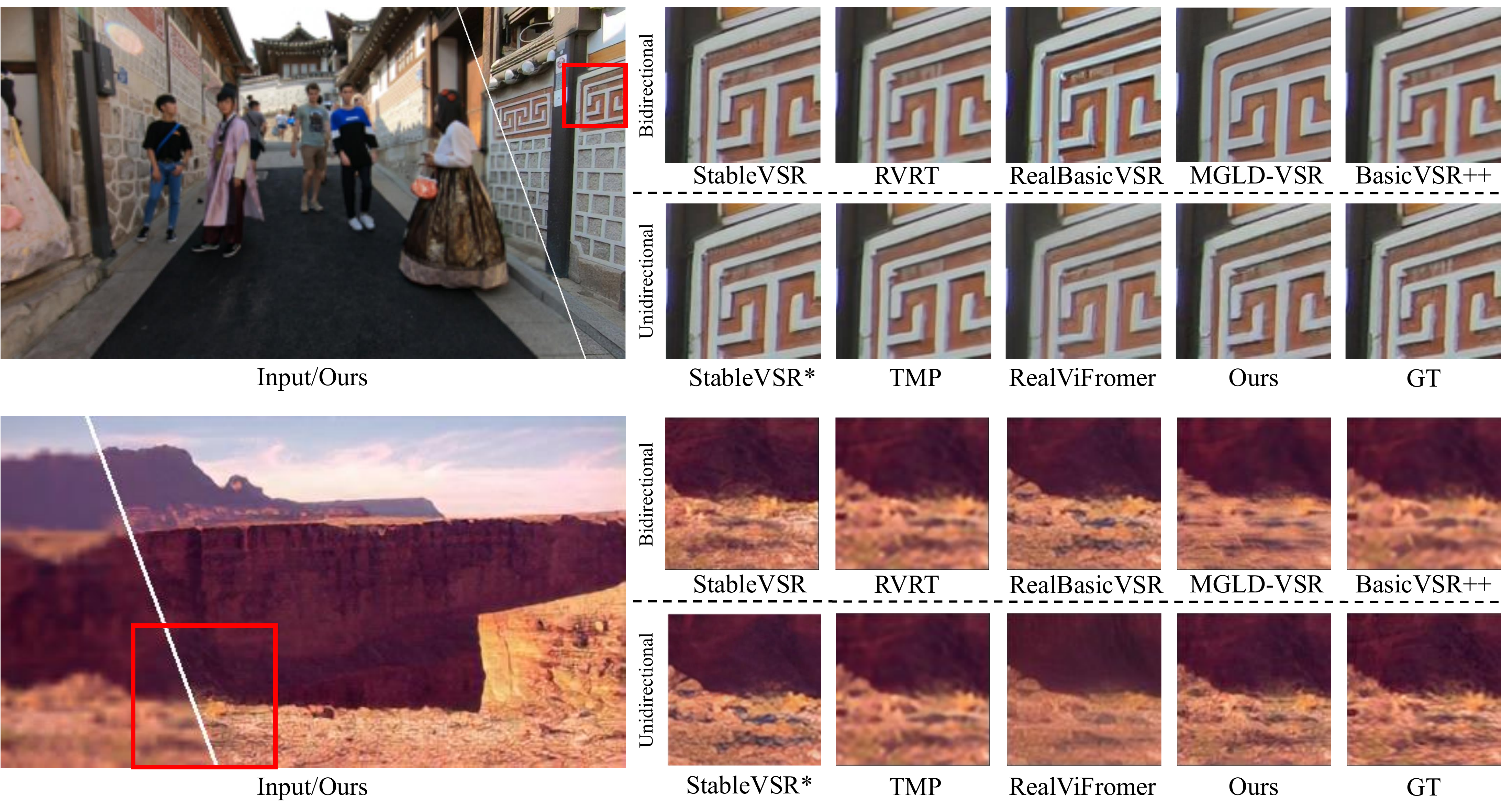}
    \vspace{-6mm}
\caption{\textbf{Qualitative comparison on REDS4 and Vimeo-90K-T datasets.} Our method demonstrates superior visual quality with sharper details compared to unidirectional methods (TMP~\cite{zhang2024tmp}, RealViformer~\cite{zhang2024realviformer}) and competitive performance against bidirectional methods (StableVSR~\cite{rota2024enhancing}, MGLD-VSR~\cite{yang2024motion}, RVRT~\cite{liang2022recurrent}, BasicVSR++\cite{chan2022basicvsr++}, RealBasicVSR\cite{chan2022investigating}).}
\label{fig:visual_result_reds4}
\end{figure*}

\vspace{-.5em}
\subsection{Qualitative Comparisons}
\vspace{-.5em}
We provide qualitative comparisons in \cref{fig:visual_result_reds4}, where Stream-DiffVSR generates sharper details and fewer artifacts than prior methods. Additional visualizations of temporal consistency and flow coherence are included in the supplemental material. A qualitative comparison with DOVE, SeedVR2, Upscale-A-Video (UAV) is included in the appendix and supplementary.

\begin{table}[t]
\caption{\textbf{Ablation study of temporal modules in Stream-DiffVSR.} 
}
\label{tab:artg_tpm_ablation}
\centering
\small
\setlength{\tabcolsep}{2pt}  
\vspace{-3mm}
\resizebox{\columnwidth}{!}{%
\begin{tabular}{lcccccccccccc}
\toprule
{Component} & LPIPS$\downarrow$ & DISTS$\downarrow$ & MUSIQ$\uparrow$ & NIQE$\downarrow$ & NRQM$\uparrow$ & BRISQUE$\downarrow$ & tLP$\downarrow$ & tOF$\downarrow$ & WarpErr$\downarrow$ & FPS$\uparrow$ & TTFF(s)$\downarrow$ \\
\midrule
Per-frame & \textcolor{red}{0.099} & 0.071 & \textcolor{blue}{65.981} & 3.249 & 6.969 & 21.655 & 7.261 & 4.201 & 25.668 & \textcolor{red}{4.807} & \textcolor{red}{0.208} \\
w/o ARTG       & 0.117 & \textcolor{blue}{0.070} & 63.347 & \textcolor{blue}{3.194} & 6.980 & 19.027 & \textcolor{blue}{6.132} & \textcolor{blue}{3.910} & \textcolor{blue}{16.598} & \textcolor{blue}{4.255} & \textcolor{blue}{0.235} \\
w/o TPM        & \textcolor{blue}{0.116} & 0.078 & \textcolor{red}{67.110} & 3.197 & 7.007 & 20.279 & 12.847 & 4.639 & 21.990 & 3.33 & 0.300 \\
TPM (unwarped) & 0.122 & 0.082 & 63.849 & 3.201 & \textcolor{blue}{7.159} & \textcolor{red}{14.063} & 12.846 & 5.689 & 17.143 & 3.125 & 0.320\\
Ours           & \textcolor{red}{0.099} & \textcolor{red}{0.062} & 65.586 & \textcolor{red}{3.111} & \textcolor{red}{7.256} & \textcolor{blue}{17.667} & \textcolor{red}{4.265} & \textcolor{red}{3.620} & \textcolor{red}{14.909} & 3.408 & 0.328 \\
\bottomrule
\end{tabular}%
}
\end{table}

\begin{table}[!htb]
\caption{\textbf{Ablation study on training strategy.}}
\label{tab:training_stage_ablation}
\setlength{\tabcolsep}{2pt}  
\vspace{-3mm}
\centering
\resizebox{\columnwidth}{!}{%
\begin{tabular}{lccccccccc}
\toprule
{Stage combination} & PSNR$\uparrow$ & SSIM$\uparrow$ & LPIPS$\downarrow$ & DISTS$\downarrow$ & MUSIQ$\uparrow$ & tLP$\downarrow$ & tOF$\downarrow$ & WarpErr$\downarrow$ \\
\midrule
stage 1 and 2 & 25.442 & 0.702 & 0.156 & 0.100 & \textcolor{red}{67.528} & 21.781 & 6.37 & 27.307 \\
stage 1 and 3 & 26.307 & 0.753 & \textcolor{blue}{0.121} & 0.077 & 64.902 & 13.094 & \textcolor{blue}{4.09} & 21.689 \\
stage 2 and 3 & \textcolor{blue}{26.906} & \textcolor{blue}{0.758} & 0.132 & 0.077 & 64.751 & \textcolor{blue}{10.510} & 4.225 & \textcolor{blue}{15.726} \\
All stage jointly & 26.135 & 0.736 & 0.124 & \textcolor{blue}{0.073} & \textcolor{blue}{67.35} & 17.816 & 4.596 & 24.298 \\
Separate (Ours) & \textcolor{red}{27.256} & \textcolor{red}{0.766} & \textcolor{red}{0.099} & \textcolor{red}{0.062} & 65.586 & \textcolor{red}{4.265} & \textcolor{red}{3.620} & \textcolor{red}{14.909} \\
\bottomrule
\end{tabular}%
}
\end{table}

\begin{table}[!htb]
\centering
\caption{\textbf{Ablation study on denoising step count within Stream-DiffVSR.} 
We evaluate 50, 10, 1, and 4 steps. Our 4-step design achieves a favorable balance between perceptual quality and runtime.}
\label{tab:denoising_steps_ablation}
\vspace{-3mm}
\setlength{\tabcolsep}{2pt}
\resizebox{\columnwidth}{!}{
\begin{tabular}{lcccccccccc}
\toprule
{Step(s)} & LPIPS$\downarrow$ & DISTS$\downarrow$ & MUSIQ$\uparrow$ & NIQE$\downarrow$ & NRQM$\uparrow$ & BRISQUE$\downarrow$ & tLP$\downarrow$ & tOF$\downarrow$ & Runtime (s)$\downarrow$ \\
\midrule
50        & \textcolor{blue}{0.102} & \textcolor{blue}{0.068} & \textcolor{red}{66.061} & \textcolor{red}{2.804} & \textcolor{blue}{7.026} & \textcolor{red}{9.925}  & 18.798 & 3.826 & 3.460 \\
10        & 0.122 & 0.072 & 64.900 & \textcolor{blue}{2.869} & 6.917 & \textcolor{blue}{12.461} & \textcolor{blue}{9.990}  & \textcolor{red}{3.625} & 0.718 \\
1         & 0.138 & 0.076 & 63.915 & 3.843 & 6.984 & 29.552 & 9.899  & 3.882 & \textcolor{red}{0.106} \\
4 (Ours)  & \textcolor{red}{0.099} & \textcolor{red}{0.062} & \textcolor{blue}{65.586} & 3.111 & \textcolor{red}{7.256} & 17.667 & \textcolor{red}{4.265} & \textcolor{blue}{3.620} & \textcolor{blue}{0.328} \\
\bottomrule
\end{tabular}
}
\end{table}

\begin{table}[!htb]
\centering
\caption{\textbf{Ablation study on Rollout Training.} 
Comparison of random timestep distillation vs. rollout training across fidelity and perceptual metrics.}
\label{tab:rollout_ablation}
\vspace{-3mm}
\setlength{\tabcolsep}{2pt}
\resizebox{\columnwidth}{!}{
\begin{tabular}{lcccccc}
\toprule
{Method} & PSNR$\uparrow$ & SSIM$\uparrow$ & LPIPS$\downarrow$ & DISTS$\downarrow$ & MUSIQ$\uparrow$ & GPU Hours$\downarrow$ \\
\midrule
Random Timestep Selection & 26.27 & 0.743 & 0.099 & 0.071 & 65.981 & 60.5 \\
Rollout Distillation      & 26.36 & 0.753 & 0.095 & 0.075 & 66.391 & 21 \\
\bottomrule
\end{tabular}
}
\end{table}

\begin{figure*}[!htb]
\begin{minipage}{.478\textwidth}
    \centering
    \includegraphics[width=\textwidth]{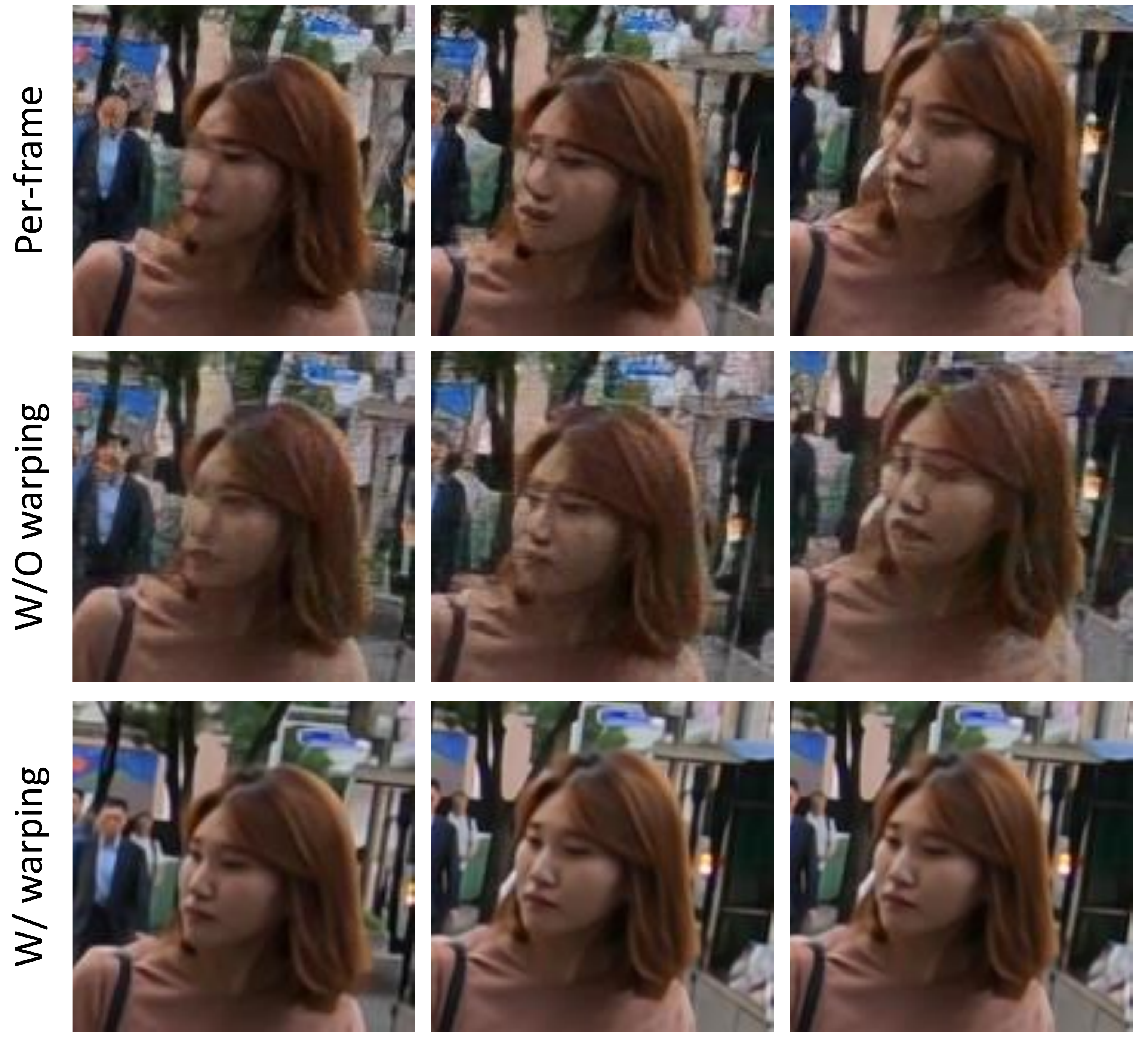}
    \vspace{-6mm}
    \caption{\textbf{Ablation study on the Temporal Processor Module (TPM).} 
    Integrating TPM improves motion stability and reduces temporal artifacts by leveraging warped previous-frame features, enhancing temporal consistency in video super-resolution.}
    \label{fig:ablation_decoder}
\end{minipage}
\hfill
\begin{minipage}{.478\textwidth}
    \centering
    \begin{minipage}{\textwidth}
        \centering
        \includegraphics[width=\textwidth]{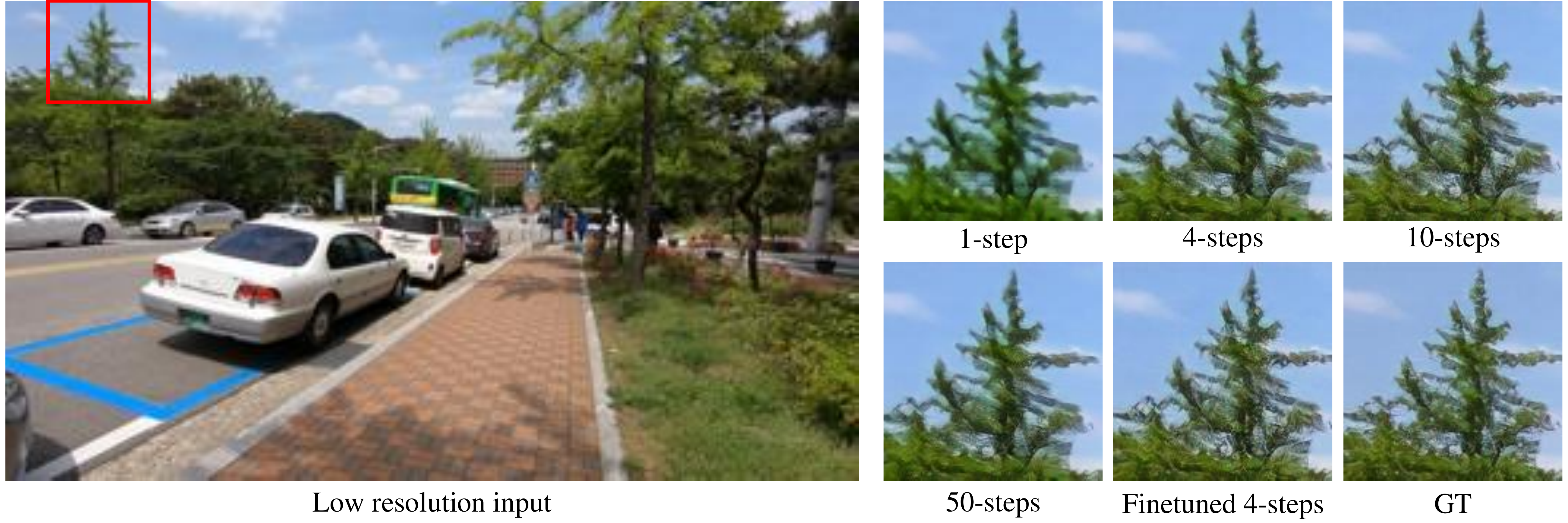}
    \vspace{-7mm}
        \caption{\textbf{Ablation study on inference steps.} 
    The 4-step model yields the best quality–efficiency trade-off, validating our distillation strategy.}
        \label{fig:ablation_unet}
    \end{minipage}
    
    
    \begin{minipage}{\textwidth}
        \centering
        \includegraphics[width=\textwidth]{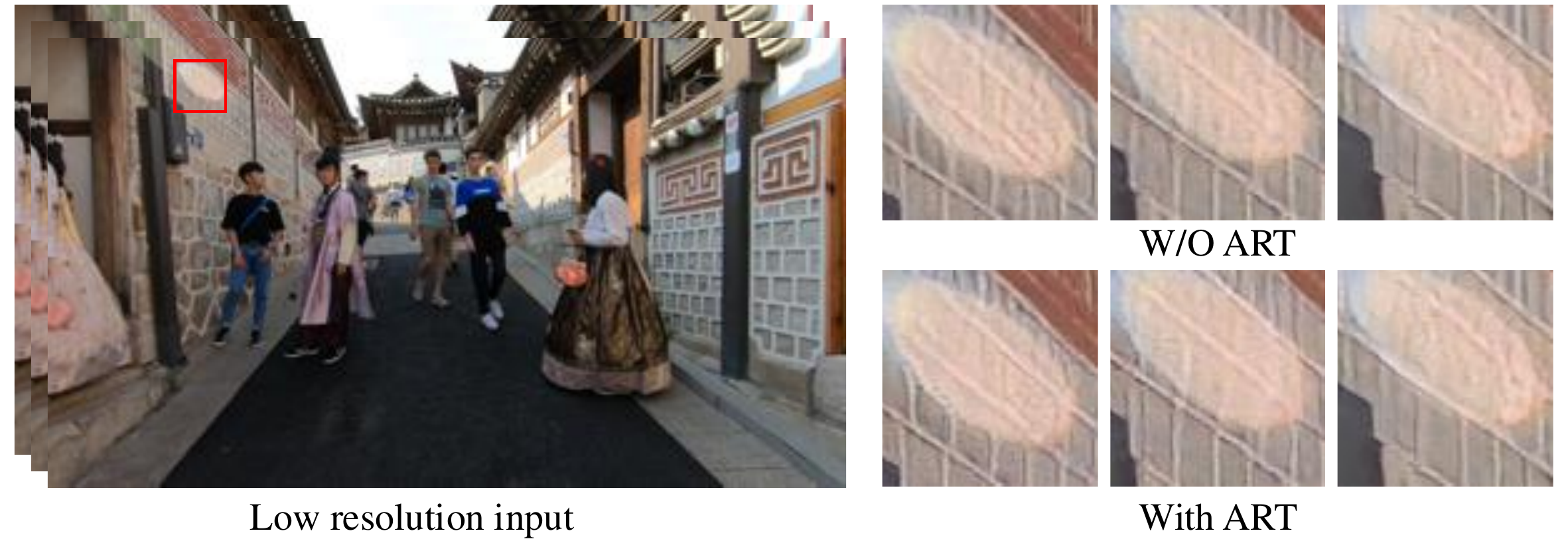}
    \vspace{-7mm}
        \caption{\textbf{Ablation study on Auto-regressive Temporal Guidance (ARTG).} 
    ARTG enhances temporal consistency and perceptual quality by leveraging warped previous frames, reducing flickering, and improving structural coherence.}
        \label{fig:ablation_ART}
    \end{minipage}
\end{minipage}
\vspace{-2mm}
\end{figure*}




\vspace{-.5em}
\subsection{Ablation Study}
\vspace{-.5em}
We ablate key components of Stream-DiffVSR including denoising-step reduction, ARTG, TPM, timestep selection, and training-stage combinations on REDS4 to ensure consistent evaluation of perceptual quality and temporal stability.

We perform ablation studies on training strategies in \cref{tab:rollout_ablation} and \cref{tab:training_stage_ablation}. For stage-wise training, partial or joint training yields inferior results, while our separate stage-wise scheme achieves the best trade-off across fidelity, perceptual, and temporal metrics. For distillation, rollout training outperforms random timestep selection in both quality and efficiency, reducing training cost from 60.5 to 21 GPU hours on 4×A6000 GPUs.

We assess the runtime–quality trade-off by varying DDIM inference steps while keeping model weights fixed. As shown in \cref{tab:denoising_steps_ablation} and \cref{fig:ablation_unet}, fewer steps increase efficiency but reduce perceptual quality, whereas more steps improve fidelity with higher latency. A 4-step setting provides the best balance.

\cref{tab:artg_tpm_ablation} and \cref{fig:ablation_ART} show the effectiveness of ARTG and TPM. The \textit{per-frame} baseline uses only the distilled U-Net with both ARTG and TPM disabled. In the ablation labels, \textit{w/o} indicates that a module is fully removed; for instance, \textit{TPM (unwarp)} feeds TPM the previous HR frame without flow-based warping, removing motion alignment. ARTG improves perceptual quality and temporal consistency. TPM further enhances temporal coherence through temporal-feature warping and fusion, yielding additional gains in tLP100. These results highlight the complementary roles of latent-space guidance and decoder-side temporal modeling.

\vspace{-.5em}
\section{Conclusion}
\label{sec:conclusion}
\vspace{-.5em}

We propose Stream-DiffVSR, an efficient online video super-resolution framework using diffusion models. By integrating a distilled U-Net, Auto-Regressive Temporal Guidance, and Temporal-aware Decoder, Stream-DiffVSR achieves superior perceptual quality, temporal consistency, and practical inference speed for low-latency applications.

\vspace{-.5em}
\paragraph{Limitations.}
Stream-DiffVSR remains heavier than CNN and Transformer models, and its use of optical flow can introduce artifacts under fast motion. Its auto-regressive design may also degrade the earliest frames, motivating stronger initialization to reduce cold-start effects. Robustness to diverse, unknown degradations remains future work.

\section*{Acknowledgements}
This research was funded by the National Science and Technology Council, Taiwan, under Grants NSTC 112-2222-E-A49-004-MY2, 113-2628-E-A49-023-, and 115-2222-E-A49-005-MY2. The authors are grateful to Google, NVIDIA, and MediaTek Inc. for their generous donations. Yu-Lun Liu acknowledges the Yushan Young Fellow Program by the MOE in Taiwan. Yu-Chih Chen acknowledges the Yushan Young Fellow Program by the MOE in Taiwan (Grant MOE-114-YSFEE-0010-008-P1).

%
%
\clearpage
\bibliographystyle{splncs04}
\bibliography{main}

\clearpage
\setcounter{page}{1}

\appendix

\vspace{-1mm}

\section*{Overview}
This supplementary material provides additional details and results to support the main paper.
We first describe the complete experimental setup in~\cref{sec:experimental_setup}, including training procedures, datasets, evaluation metrics, and baseline configurations. We then present extended implementation details and a three-stage breakdown of our training pipeline in~\cref{sec:implementation_details,sec:training_details}, covering U-Net distillation, temporal-aware decoder training, and the Auto-regressive Temporal Guidance module.
Next, we report additional visual comparisons on multiple benchmarks under both bidirectional and unidirectional settings in~\cref{sec:additional_vis}, followed by extensive qualitative visualizations illustrating perceptual quality, temporal consistency. And comparison with FlashVSR~\cite{zhuang2025flashvsr} in ~\cref{sec:suppl_flashvsr_comparison}. We also include representative failure cases to highlight current limitations in~\cref{sec:failure_cases}.
\noindent\textbf{Video results.} We provide an accompanying HTML index for interactive playback of all supplementary videos and zoom-in comparisons.

\section{Experimental Setup} \label{sec:experimental_setup}


\subsection{Training and Evaluation Setup} Stream-DiffVSR is trained in three sequential stages to ensure stable optimization and modular control over temporal components. All evaluation experiments are conducted on an NVIDIA RTX 4090 GPU with TensorRT acceleration, unless otherwise specified.
In particular, the resource-focused comparisons (runtime/TTFF/peak memory) against memory-intensive baselines are measured on a single NVIDIA RTX Pro 6000 GPU, as reported in the corresponding tables. Details of the stage-wise training procedure and configurations are provided in the supplementary.
\paragraph{Runtime and Time-to-First-Frame measurement.}
All reported runtime and TTFF numbers are measured end-to-end on the GPU, \emph{including} optical-flow estimation and all model components used at inference time (temporal modules and post-processing).
We use a warm-up run and report the average per-frame runtime over the evaluation sequence; \emph{Time-to-First-Frame (TTFF)} denotes the maximum end-to-end output delay over the sequence.

\paragraph{Resolution alignment for evaluation.}
Some diffusion Transformer (DiT)-based baselines produce outputs at a fixed resolution that may not match the ground-truth (GT) frame size required by our benchmarks. 
To ensure a fair and consistent evaluation across all methods, we align the output resolution to the GT resolution before computing metrics. 
Specifically, when a method’s native output resolution differs from the GT, we apply bicubic downsampling to resize the restored frames to the GT resolution; otherwise, no resizing is performed. 
All full-reference (e.g., PSNR/SSIM/LPIPS/DISTS/VMAF/tLP/tOF) metrics are computed on these resolution-aligned outputs.

\subsection{Datasets}
We evaluate our method using widely-recognized benchmarks: REDS~\cite{Nah_2019_CVPR_REDS} and Vimeo-90K~\cite{xue2019video}. REDS consists of 300 video sequences (1280$\times$720 resolution, 100 frames each); sequences 000, 011, 015, and 020 (REDS4) are used for testing. Vimeo-90K-T contains 91,701 clips (448$\times$256 resolution), with 64,612 for training and 7,824 (Vimeo-90K) for evaluation, offering diverse content for training.

For testing under complex degradation, we also evaluate on two additional benchmarks: VideoLQ~\cite{chan2022investigating}, a no-reference video quality dataset curated from real Internet content, and Vid4~\cite{liu2013bayesian}, a classical benchmark with 4 videos commonly used for VSR evaluation. The evaluation results are provided in supplementary.


\subsection{Evaluation metrics}
We assess the effectiveness of our approach using a comprehensive set of perceptual and temporal metrics across multiple aspects.
\textbf{Reference-based Perceptual Quality}: LPIPS~\cite{zhang2018unreasonable} and DISTS~\cite{ding2020image}.
\textbf{No-reference Perceptual Quality}: MUSIQ~\cite{ke2021musiq}, NIQE~\cite{saad2012blind}, NRQM~\cite{ma2017learning}, BRISQUE~\cite{mittal2012no}.
\textbf{Temporal Consistency}: Temporal Learned Perceptual Similarity (tLP), and Temporal Optical Flow difference (tOF).
\textbf{Inference Speed}: Per-frame runtime, Time-to-First-Frame (TTFF) measured on an NVIDIA RTX 4090 GPU to evaluate low-latency applicability.
Note that while we report PSNR and SSIM results (REDS4: 27.256 / 0.768) for completeness, we do not rely on these distortion-based metrics in our main analysis, as they often fail to reflect perceptual quality and temporal coherence, especially in generative VSR settings. This has also been observed in prior work~\cite{zhang2018unreasonable}. Our qualitative results demonstrate superior perceptual and temporal quality, as we prioritize low-latency stability and consistency over overfitting to any single metric.

\subsection{Baseline methods}
We evaluate our method against leading CNN-based, Transformer-based, and Diffusion-based models. Specifically, we include bidirectional (offline) methods such as BasicVSR++~\cite{chan2022basicvsr++}, RealBasicVSR~\cite{chan2022investigating}, RVRT~\cite{liang2022recurrent}, StableVSR~\cite{rota2024enhancing}, MGLD-VSR~\cite{yang2024motion}, and unidirectional (online) methods including MIA-VSR~\cite{zhou2024video}, TMP~\cite{zhang2024tmp}, RealViformer~\cite{zhang2024realviformer}, and StableVSR$^*$~\cite{rota2024enhancing}, comprehensively comparing runtime, perceptual quality, and temporal consistency. 
In addition, we consider several memory-intensive diffusion/DiT-style and long-video enhancement baselines, including VEnhancer~\cite{he2024venhancer}, Upscale-A-Video~\cite{zhou2024upscale}, SeedVR2~\cite{wang2025seedvr2}, DOVE~\cite{zhang2025dove}, and FlashVSR~\cite{zhuang2025flashvsr} (as well as its lightweight variant FlashVSR-tiny~\cite{zhuang2025flashvsr}), and report their runtime/TTFF and peak GPU memory usage where applicable. For FlashVSR, we follow the official implementation and evaluation setting, including the temporal chunk size of 6; we report TTFF and peak GPU memory under this configuration. Unless otherwise noted, all baselines are evaluated using official implementations with the authors' recommended settings.\footnotetext{$^*$StableVSR~\cite{rota2024enhancing} is originally a bidirectional model. We implement a unidirectional variant (StableVSR$^*$) that only uses forward optical flow for fair comparison under the \hl{unidirectional} setting.}

\section{Additional Implementation Details} \label{sec:implementation_details}
\subsection{Implementation Details}Our UNet backbone is initialized from the StableVSR~\cite{rota2024enhancing} released UNet checkpoint, which is trained for image-based super-resolution from Stable Diffusion (SD) x4 Upscaler~\cite{rombach2022high,Rombach_2022_CVPR}. We then perform 4-step distillation to adapt this UNet for efficient video SR. ARTG, in contrast, is built upon our distilled UNet encoder and computes temporal residuals from previous high-resolution outputs using convolutional and transformer blocks. These residuals are injected into the decoder during upsampling, enhancing temporal consistency without modifying the encoder or increasing diffusion steps. Our decoder is initialized from AutoEncoderTiny and extended with a Temporal Processor Module (TPM) to incorporate multi-scale temporal fusion during final reconstruction.

\section{Additional Training Detials} \label{sec:training_details}
\subsection{Stage 1: U-Net Distillation} We initialize the denoising U-Net from the 50-step diffusion model released by StableVSR~\cite{rota2024enhancing}, which was trained on REDS~\cite{Nah_2019_CVPR_REDS} dataset. Both the released StableVSR checkpoint and our subsequent training use only the REDS \emph{training} split, following the same convention as BasicVSR++~\cite{chan2022basicvsr++} and MIA-VSR~\cite{zhou2024video}; the REDS4 sequences used for evaluation are never seen during training. To accelerate inference, we distill the 50-step U-Net into a 4-step variant using a deterministic DDIM~\cite{song2020denoising} scheduler. During training, our rollout distillation always starts from the noisiest latent 
at timestep $999$ and executes the full sequence of four denoising steps 
$\{999 , 749 , 499 , 249\}$. 
Supervision is applied only to the final denoised latent at $t=0$, ensuring 
that training strictly mirrors the inference trajectory and reducing 
the gap between training and inference. We use a batch size of 16, learning rate of 5e-5 with constant, and AdamW optimizer ($\beta_1=0.9$, $\beta_2=0.999$, weight decay 0.01). Training is conducted for 600K iterations with a patch size of $512\times512$.The distillation loss consists of MSE loss in latent space, LPIPS~\cite{zhang2018unreasonable} loss, and adversarial loss using a PatchGAN discriminator~\cite{isola2017image} in pixel level, with weights of 1.0, 0.5, and 0.025 respectively. Adversarial loss are envolved after 20k iteration for training stabilization.

\subsection{Stage 2: Temporal-aware Decoder Training} The decoder receives both the encoded ground truth latent features and temporally aligned context features (via flow-warped previous frames). The encoder used to extract temporal features is frozen.We use a batch size of 16, learning rate of 5e-5 with constant, and AdamW optimizer ($\beta_1=0.9$, $\beta_2=0.999$, weight decay 0.01). Training is conducted for 600K iterations with a patch size of $512\times512$. Loss consists of smooth L1 reconstruction loss, LPIPS~\cite{zhang2018unreasonable} loss, flow loss using RAFT~\cite{teed2020raft} and adversarial loss using a PatchGAN discriminator~\cite{isola2017image} in pixel level for training, with weights of 1.0, 0.3, 0.1 and 0.025 respectively. Flow loss and adversarial loss are envolved after 20k iteration for training stabilization.

\subsection{Stage 3: Auto-regressive Temporal Guidance} We train the ARTG module while freezing both the U-Net and decoder. Optical flow is computed between adjacent frames using RAFT~\cite{teed2020raft}, and the warped previous super-resolved frame is injected into the denoising U-Net and decoder. The loss formulation is identical to Stage 1, conducted with 60K iterations. This guides ARTG to enhance temporal coherence while maintaining alignment with the original perceptual objectives.

\begin{algorithm}[h]
\caption{Training procedure for U-Net rollout distillation.}
\label{alg:unet_distill}
\footnotesize
\KwIn{Dataset \(\mathcal{D}=\{(\tilde{I}, I)\}\); pre-trained VAE; 4-step noise scheduler; student U-Net with parameters \(\theta\); discriminator \(D(\cdot)\).}
\For{epoch \(=1\) to \(N\)}{
  \For{each batch \((\tilde{I}, I) \in \mathcal{D}\)}{
    \(\mathbf{z}_0 \gets \mathrm{VAE.encode}(I)\)\;
    Sample \(\epsilon \sim \mathcal{N}(0,I)\)\;
    \(\mathbf{z}_T \gets \alpha_T \mathbf{z}_0 + \sqrt{1-\alpha_T}\,\epsilon\) \tcp*{Add noise at maximum timestep $T$}
    
    \tcp{--- Rollout 4-step denoising ---}
    \(\hat{\mathbf{z}}_T \gets [\,\mathbf{z}_T,\,\tilde{I}]\)\;
    \For{step $s = T, \dots, 1$}{
        \(\hat{\epsilon} \gets \mathrm{U\!-\!Net}(\hat{\mathbf{z}}_s, s)\)\;
        \(\hat{\mathbf{z}}_{s-1} \gets \mathrm{Scheduler.step}(\hat{\epsilon}, s, \hat{\mathbf{z}}_s)\)\;
    }
    
    \(\hat{I} \gets \mathrm{VAE.decode}(\hat{\mathbf{z}}_0)\)\;
    
    \(\mathcal{L}_{\mathrm{L2}} \gets \|\hat{I} - I\|_2^2\)\;
    \(\mathcal{L}_{\mathrm{LPIPS}} \gets \mathrm{LPIPS}(\hat{I}, I)\)\;
    \(\mathcal{L}_{\mathrm{GAN}} \gets \mathrm{softplus}\bigl(-D(\hat{I})\bigr)\)\;
    \(\mathcal{L} \gets \lambda_{L2}\,\mathcal{L}_{\mathrm{L2}} + \lambda_{\mathrm{LPIPS}}\,\mathcal{L}_{\mathrm{LPIPS}} + \lambda_{\mathrm{GAN}}\,\mathcal{L}_{\mathrm{GAN}}\)\;
    
    Update parameters: \(\theta \gets \theta - \eta \nabla_{\theta}\mathcal{L}\)\;
  }
}
\end{algorithm}

\begin{algorithm}[h]
\SetAlgoNoLine
\SetAlgoNoEnd
\caption{Auto-Regressive Diffusion VSR.}
\label{alg:ar_diff_vsr}
\footnotesize
\textbf{Notation:}
\(\{\tilde{I}_i\}\): Input LR frames, \(\{\hat{I}_i\}\): Enhanced frames, \(\mathrm{FlowWarp}\): Warping w.r.t. flow, \(\mathrm{VAE}\): Auto-regressive VAE, \(\mathrm{U\!Net}\): Distilled diffusion U-Net, \(\mathrm{ARTG}\): Auto-Regressive Temporal guidance, \(\mathrm{PrepareLatents}\): Create latent input, \(\mathrm{timesteps}\): \(\{t_1,\dots,t_4\}\)

\textbf{Input:} \(\{\tilde{I}_i\}_{i=1}^N\), flows \(\{\mathbf{f}_{i-1}\}_{i=2}^N\), \(\mathrm{VAE}, \mathrm{U\!Net}, \mathrm{ARTG}\).\\
\textbf{Output:} \(\{\hat{I}_i\}_{i=1}^N\).\\[4pt]

\For{$i=1$ \KwTo $N$}{
  $\mathbf{LQ}_i \gets \tilde{I}_i$\\
  $\mathbf{z}_i \gets \mathrm{PrepareLatents}(\mathbf{LQ}_i,\,t)$\\
  \If{$i>1$}{
    $\hat{I}_{i-1}^{w} \gets \mathrm{FlowWarp}(\hat{I}_{i-1},\,\mathbf{f}_{i-1})$\\
    $\mathbf{E}_{i-1} \gets \mathrm{VAE.encode}(\hat{I}_{i-1}^{w})$\\
  }
  \For{$t \in \mathrm{timesteps}$}{
    \If{$i>1$}{
      $\mathbf{z}_i \gets \mathrm{ARTG}(\mathbf{z}_i,\,\hat{I}_{i-1}^{w})$\\
    }
    $\hat{\epsilon} \gets \mathrm{U\!Net}(\mathbf{z}_i,\,t)$\\
    $\mathbf{z}_i \gets \text{DiffusionUpdate}(\hat{\epsilon},\,t,\,\mathbf{z}_i)$\\
  }
  \If{$i>1$}{
    $\hat{I}_i \gets \mathrm{VAE.Decode}(\mathbf{z},\,\mathbf{E}_{i-1})$\\
  }
  \Else{
    $\hat{I}_i \gets \mathrm{VAE.Decode}(\mathbf{z})$\\
  }
}
\Return{$\{\hat{I}_i\}$}
\end{algorithm}

\section{Additional Visual Result} \label{sec:additional_vis}
\begin{figure*}[t]
    \centering
    \includegraphics[width=0.8\linewidth]{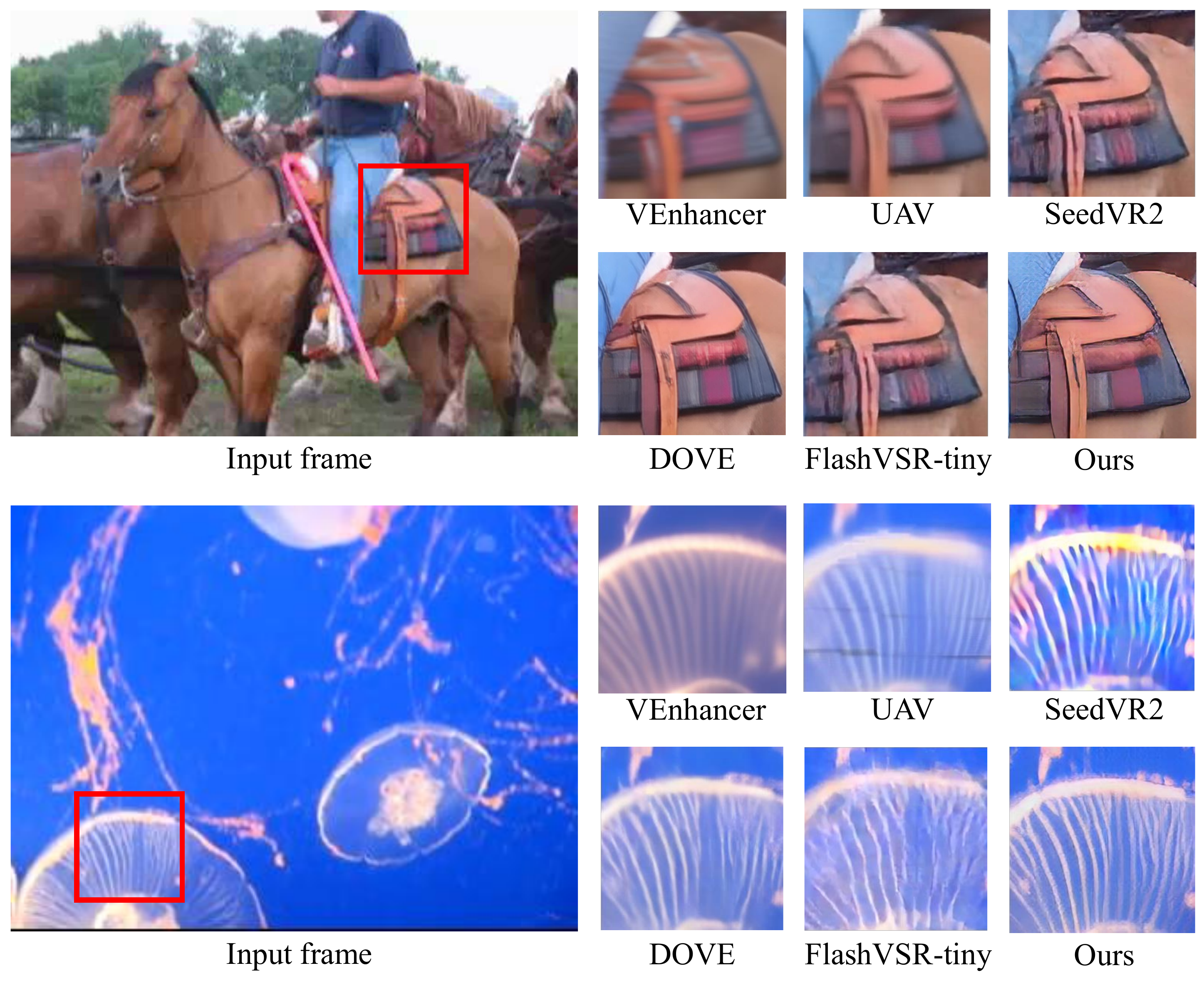}
    \caption{\textbf{Additional visual results on VideoLQ dataset.}}
    \label{fig:suppl_v4}
\end{figure*}

\begin{figure*}[t]
    \centering
    \includegraphics[width=0.8\linewidth]{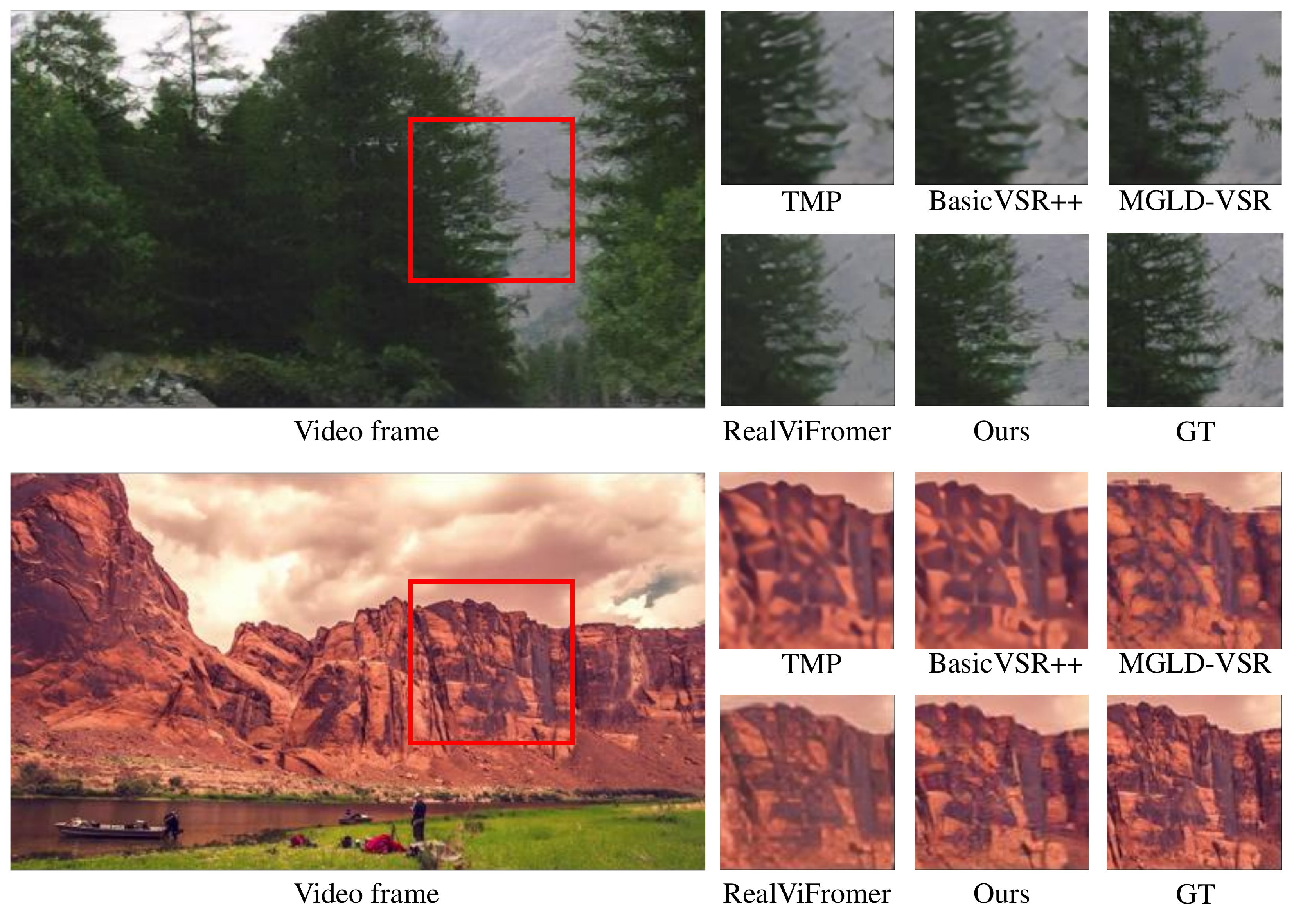}
    \caption{\textbf{Additional visual results on Vimeo-90K-T dataset.}}
    \label{fig:suppl_v1}
\end{figure*}

\begin{figure*}[t]
    \centering
    \includegraphics[width=0.8\linewidth]{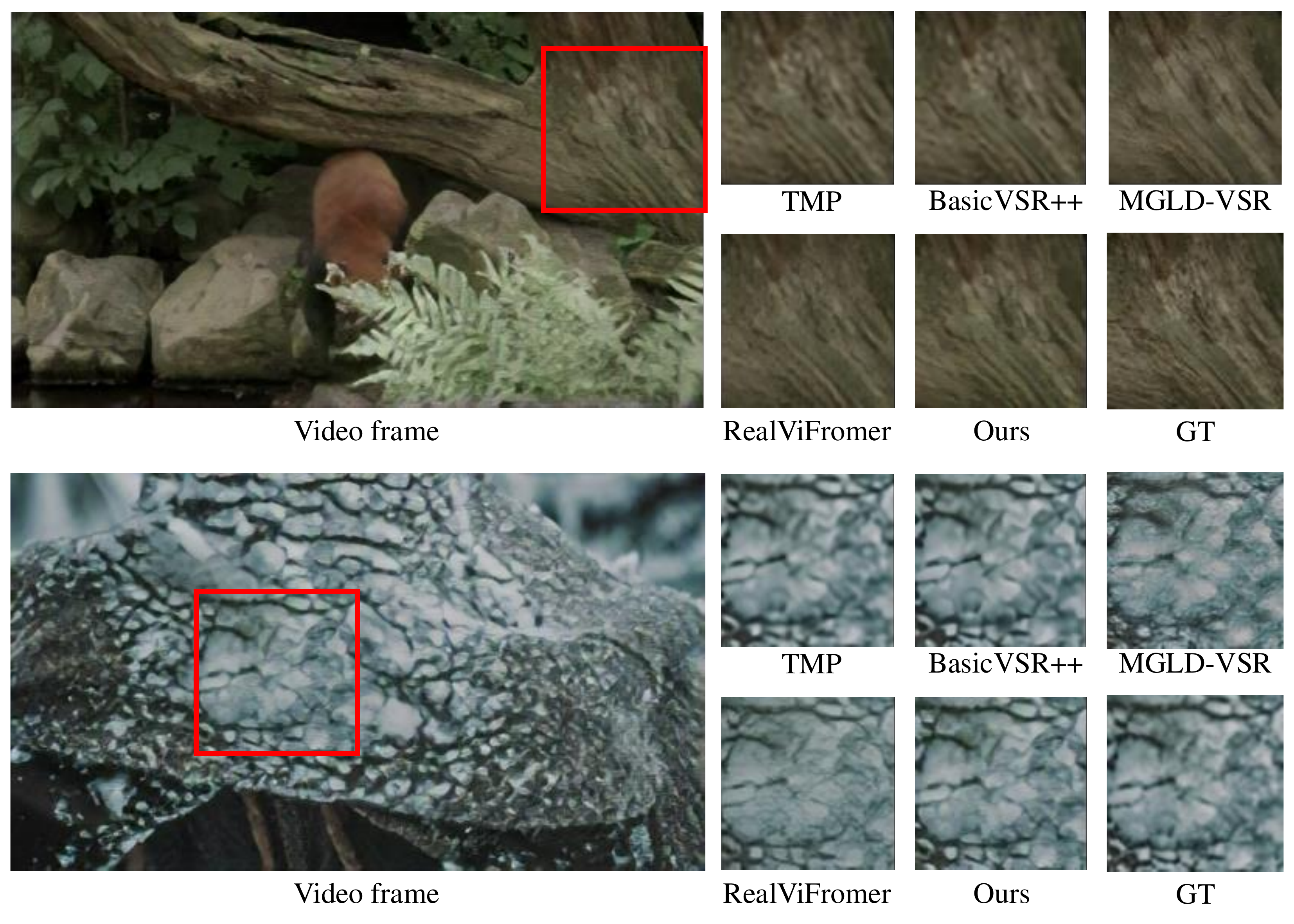}
    \caption{\textbf{Additional visual results on Vimeo-90K-T dataset.}}
    \label{fig:suppl_v2}
\end{figure*}

\begin{figure*}[t]
    \centering
    \includegraphics[width=0.8\linewidth]{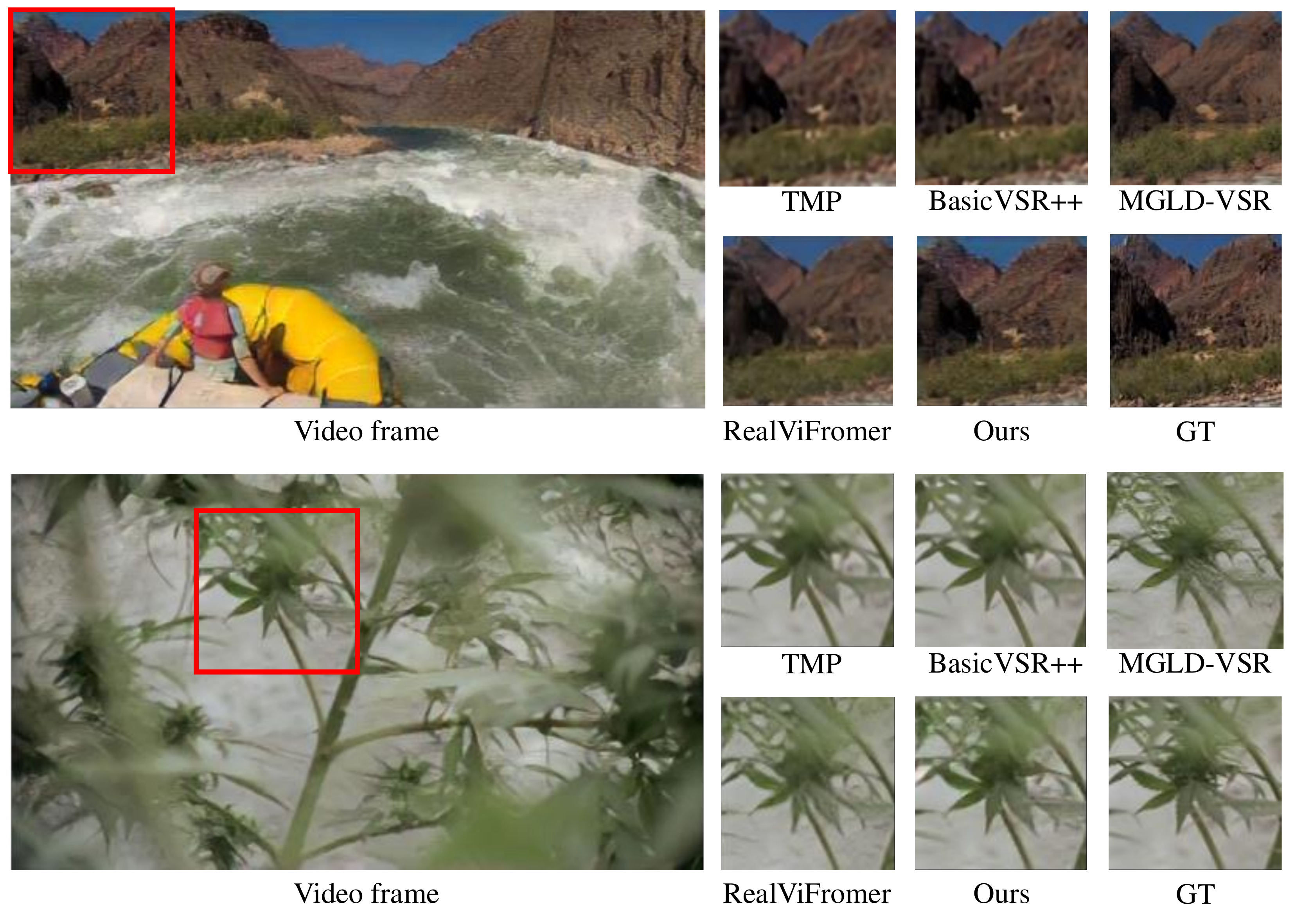}
    \caption{\textbf{Additional visual results on Vimeo-90K-T dataset.}}
    \label{fig:suppl_v3}
\end{figure*}
\begin{figure*}[t]
    \centering
    \includegraphics[width=\textwidth]{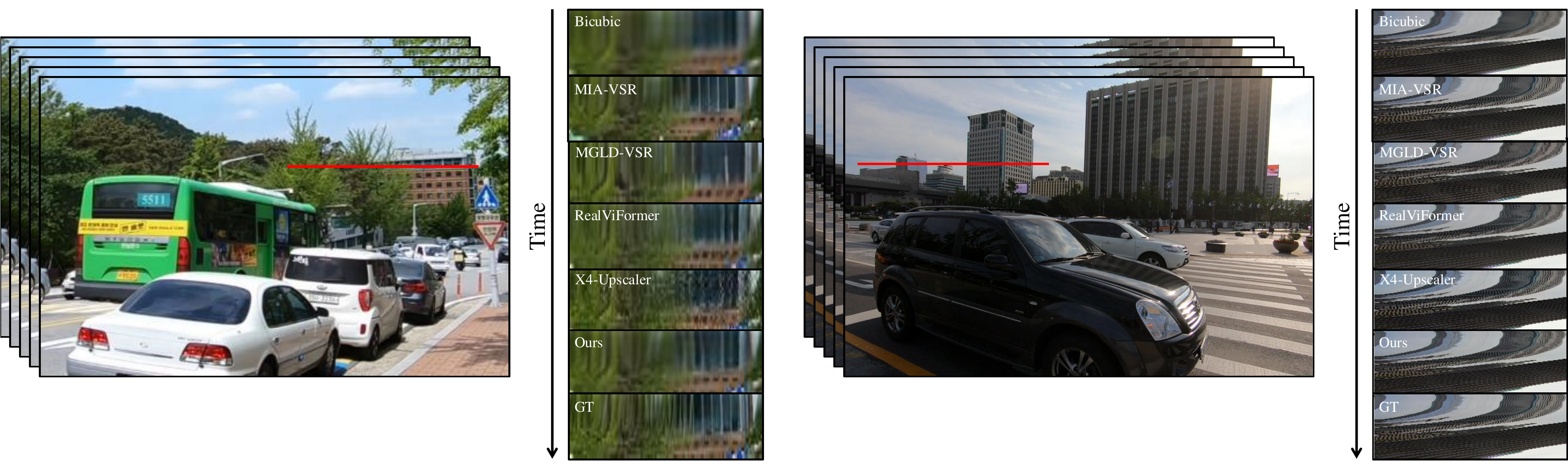}
\vspace{-7mm}
    \caption{\textbf{Temporal consistency comparison.} Qualitative comparison of temporal consistency across consecutive frames. Our proposed Stream-DiffVSR effectively mitigates flickering artifacts and maintains stable texture reconstruction, demonstrating superior temporal coherence compared to existing VSR methods.}
    \label{fig:temporal_vis}
\end{figure*}

\begin{figure*}[h]
    \centering
    \includegraphics[width=\columnwidth,height=0.5\columnwidth,keepaspectratio]{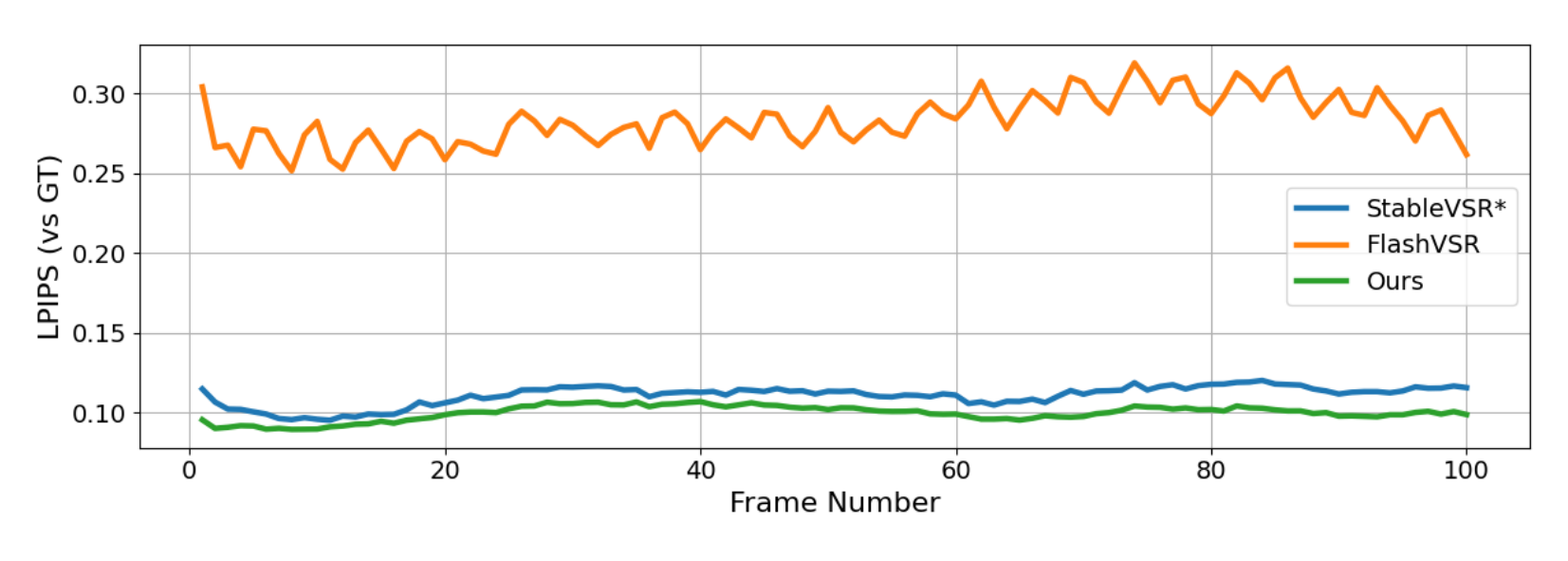}
    \vspace{-3em}
    \caption{\textbf{Per-frame LPIPS comparison on REDS4 dataset.}}
    \label{fig:lpips_comparison}
\end{figure*}
\begin{figure*}[t]
    \centering
    \includegraphics[width=\textwidth]{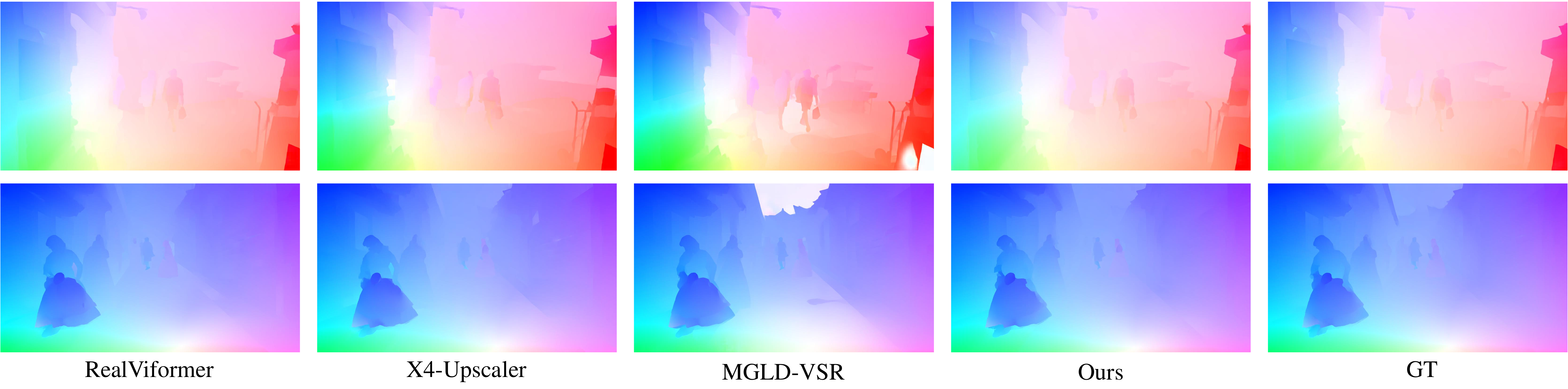}
\vspace{-6mm}
    \caption{\textbf{Optical flow visualization comparison.} Visualization of optical flow consistency across different VSR methods. Our proposed Stream-DiffVSR produces smoother and more temporally coherent flow fields, indicating improved motion consistency and reduced temporal artifacts compared to competing approaches.}
    \label{fig:flow_vis}
\end{figure*}

\begin{figure*}[t]
    \centering
    \includegraphics[width=\linewidth]{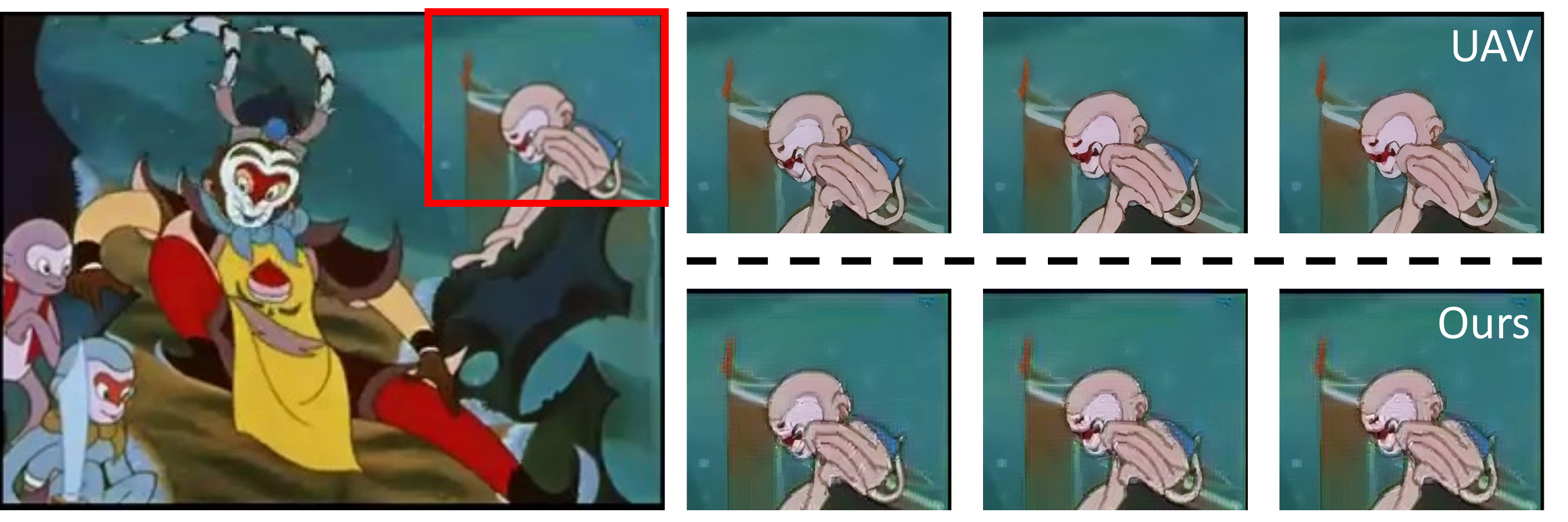}
\vspace{-3mm}
\caption{\textbf{Qualitative comparison with Upscale-A-Video (UAV) on AIGC video frames.}}

\label{fig:uav_comparison}
\end{figure*}

~\cref{fig:suppl_v1,fig:suppl_v2,fig:suppl_v3,fig:suppl_v4} presents qualitative results on challenging sequences with diverse content and motion. We also provide qualitative comparisons with Upscale-A-Video~\cite{zhou2024upscale} on AIGC video frame in~\cref{fig:uav_comparison}. Compared with CNN-based (TMP~\cite{zhang2024tmp}, BasicVSR++~\cite{chan2022basicvsr++}) and Transformer-based (RealViformer~\cite{zhang2024realviformer}) approaches, as well as diffusion-based methods (e.g., MGLD-VSR) and several memory-intensive baselines (e.g., VEnhancer~\cite{he2024venhancer}, SeedVR2~\cite{wang2025seedvr2}, DOVE~\cite{zhang2025dove}, Upscale-A-Video and FlashVSR~\cite{zhuang2025flashvsr} where available), our method produces sharper structures and more faithful textures. These visual comparisons further demonstrate the effectiveness of our design in maintaining perceptual quality and temporal consistency across diverse scenes.

\textbf{Temporal consistency comparison.} As shown in the consecutive-frame comparisons~\cref{fig:temporal_vis}, Stream-DiffVSR alleviates flickering artifacts and preserves stable textures over time, yielding noticeably stronger temporal coherence than prior VSR methods. To verify that the staged training and rollout distillation described in the main paper keep error accumulation controllable over long sequences, we plot LPIPS as a function of frame index over 100-frame clips from REDS4. \cref{fig:lpips_comparison} compares Stream-DiffVSR against StableVSR* and FlashVSR. Stream-DiffVSR maintains the lowest LPIPS throughout the sequence with no upward drift, whereas StableVSR* shows a clear increase in LPIPS over time, and FlashVSR exhibits visible jitter consistent with its chunk-wise processing.

\textbf{Optical flow visualization comparison.} The optical flow consistency visualizations~\cref{fig:flow_vis} further highlight our advantages: Stream-DiffVSR generates smoother and more temporally coherent flow fields, reflecting improved motion stability and reduced temporal artifacts.


\section{Comparison with FlashVSR / FlashVSR-Tiny}
\label{sec:suppl_flashvsr_comparison}

As discussed in the main paper, chunk-based methods such as FlashVSR must buffer $k$ input frames before producing any output, so their Time-to-First-Frame (TTFF) scales as $k \times T_\text{frame}$, whereas Stream-DiffVSR processes frames one at a time and achieves $\text{TTFF} = T_\text{frame}$. \cref{tab:suppl_flashvsr_chunk} reports the chunk size used by each method alongside runtime, TTFF, and peak memory on a single NVIDIA RTX Pro 6000 GPU.

\begin{table}[h]
\centering\small
\caption{\textbf{Chunk size and runtime comparison with FlashVSR / FlashVSR-Tiny on REDS4.} Runtime, TTFF, and peak memory are measured on a single NVIDIA RTX Pro 6000 GPU. Best and second-best results are marked in \textcolor{red}{red} and \textcolor{blue}{blue}.}
\label{tab:suppl_flashvsr_chunk}
\begin{tabular}{lccccc}
\toprule
Method & Chunk size ($k$) & Runtime (s)$\downarrow$ & TTFF (s)$\downarrow$ & Mem (GB)$\downarrow$ \\
\midrule
FlashVSR & 6 & 0.283 & 1.698 & 19.630 \\
FlashVSR-Tiny & 6 & \textcolor{red}{0.157} & \textcolor{blue}{0.942} & \textcolor{red}{12.390} \\
Stream-DiffVSR (ours) & 1 & \textcolor{blue}{0.243} & \textcolor{red}{0.243} & \textcolor{blue}{19.580} \\
\bottomrule
\end{tabular}
\end{table}

\begin{figure}[h]
    \centering
    \includegraphics[width=\columnwidth,height=0.7\columnwidth,keepaspectratio]{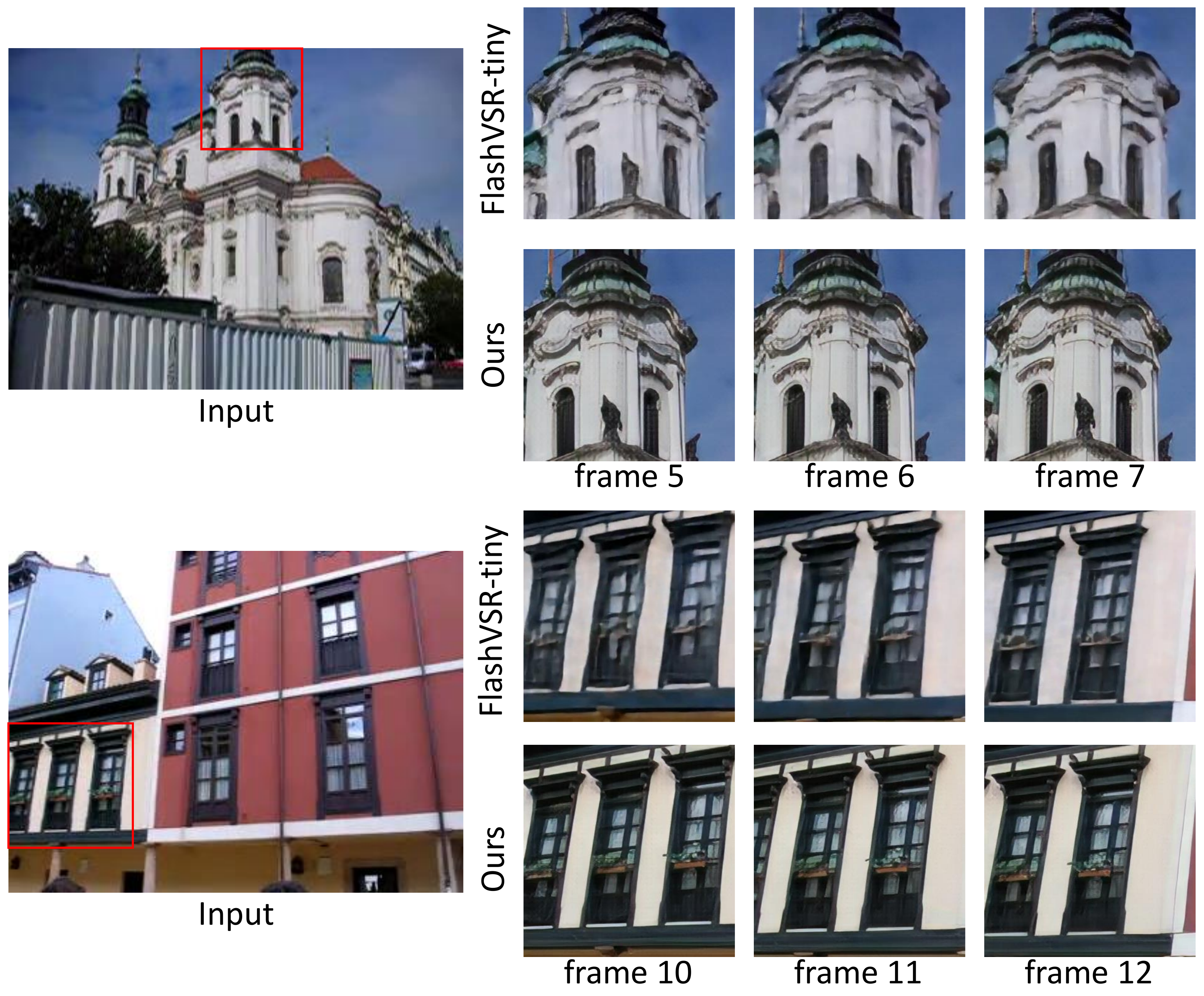}
    \caption{\textbf{Additional visual comparison on VideoLQ dataset.}}
    \label{fig:flashvsr_ours_comparison}
    \vspace{-4mm}
\end{figure}

We additionally provide qualitative comparisons against FlashVSR-Tiny on the real-world VideoLQ benchmark, using the mixed-degradation fine-tuned checkpoint introduced in \cref{sec:mixed_degradation}; note that the original FlashVSR encounters out-of-memory failure under this setting. \cref{fig:flashvsr_ours_comparison} shows two additional cases covering fine textures and fast motion.

\section{Failure cases} \label{sec:failure_cases}
\begin{figure}[h]
    \centering
    \includegraphics[width=\columnwidth]{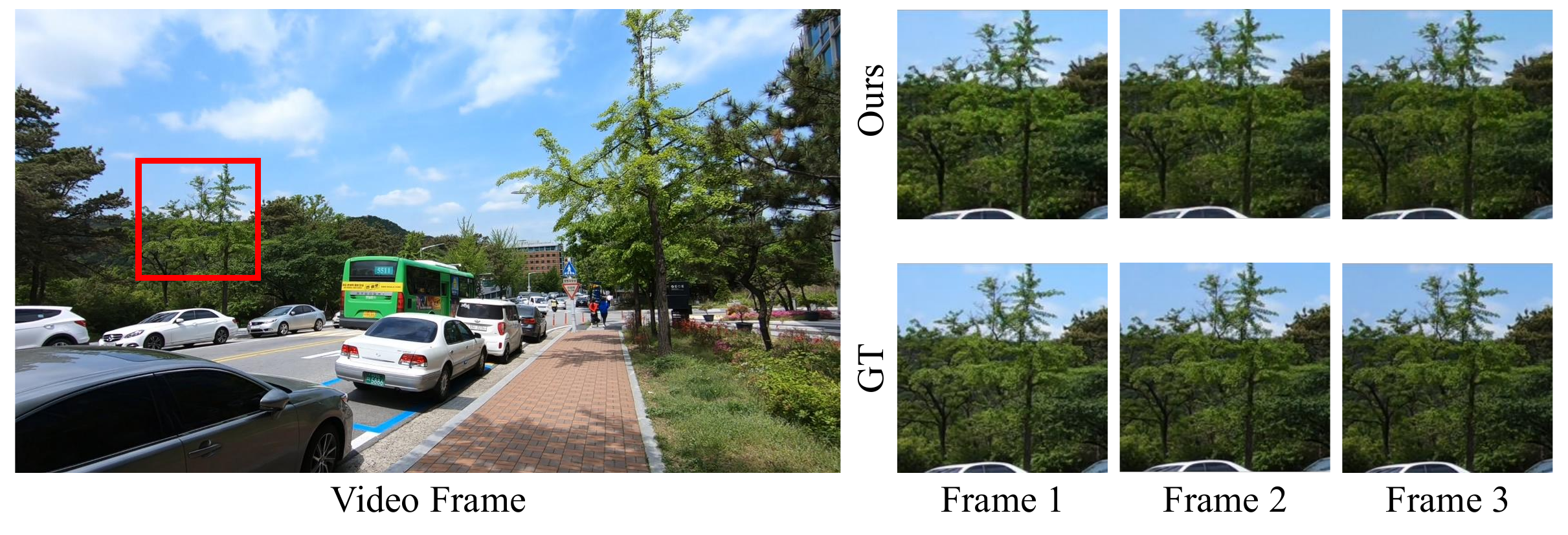}
    \vspace{-6mm}
    \caption{
    \textbf{Limitation on the first frame without temporal context.}
   Our method may underperform on the first frame of a video sequence due to the absence of prior temporal information. This limitation is inherent to online VSR settings, where no past frames are available for guidance.}
    \label{fig:failure_cases}
\end{figure}

~\cref{fig:failure_cases} illustrates a limitation of our approach on the first frame of a video sequence. 
Since no past frames are available for temporal guidance, the model may produce blurrier details or less stable structures compared to subsequent frames. 
This issue is inherent to all online VSR settings, where temporal information cannot be exploited at the sequence start. 
As shown in later frames, once temporal context becomes available, our method quickly stabilizes and reconstructs high-fidelity details.



\end{document}